\documentclass[10pt,twocolumn,letterpaper]{article}

\usepackage{cvpr}              
\usepackage{amsmath}
\usepackage{amssymb}
\usepackage{pifont}
\usepackage{multirow}
\usepackage{comment}

\newcommand{\xmark}{\ding{55}}%
\definecolor{cvprblue}{rgb}{0.21,0.49,0.74}
\usepackage[pagebackref,breaklinks,colorlinks,allcolors=cvprblue]{hyperref}

\newcommand\blfootnote[1]{%
  \begingroup
  \renewcommand\thefootnote{}\footnote{#1}%
  \addtocounter{footnote}{-1}%
  \endgroup
}

\def\paperID{6} 
\def\confName{CVPR}
\def\confYear{2025}

\title{Leveraging Automatic CAD Annotations for \\Supervised Learning in 3D Scene Understanding}

\author{Yuchen Rao$^{1}$\textsuperscript{*}
\quad
Stefan Ainetter$^{1}$\textsuperscript{*}
\quad
Sinisa Stekovic$^{2}$
\quad
 Vincent Lepetit$^{2}$
\quad
Friedrich Fraundorfer$^{1}$
\and
{\normalsize $^1$ Inst. of Visual Computing, Graz Univ. of Technology, Austria} \\ {\normalsize
$^2$ LIGM, \'Ecole des Ponts et Chaussees, IP Paris, CNRS, France}
\and
{\tt\small Project page: \href{https://stefan-ainetter.github.io/SCANnotatepp/}{stefan-ainetter.github.io/SCANnotatepp}}
}

\usepackage{dsfont}

\newif\ifshowedits

\newcommand{\addeditor}[3]{%
  \definecolor{#1color}{rgb}{#3}
  \expandafter\newcommand\csname #1\endcsname[1]{
  \ifshowedits
    {\color{#1color} ##1}
  \else
    {##1}
  \fi
  }%
  \expandafter\newcommand\csname #1rmk\endcsname[1]{
  \ifshowedits
    {\color{#1color} {\bf [#2: ##1]}}
  \else
    {}
  \fi
  }%
}

\newcommand{\createtextvar}[1]{
  \expandafter\newcommand\csname #1\endcsname{%
  {\text{#1}}
}%
}
\newcommand{\textvars}[1]{\forcsvlist{\createtextvar}{#1}}

\newcommand{\calL}{{\cal L}}

\addeditor{sinisa}{SS}{1.0, 0.0, 0.0}
\addeditor{yuchen}{YR}{0.8, 0.0, 0.8}
\addeditor{stefan}{SA}{0.0, 0.0, 0.8}
\addeditor{friedrich}{FF}{0.0, 0.0, 0.5}
\addeditor{vincent}{VL}{0, 0.6, 0}
\showeditstrue

\begin{document}
\twocolumn[{
    \renewcommand\twocolumn[1][]{#1}   
    \maketitle
    \centering 
    \scalebox{0.9}{
    \begin{tabular}{cc}
        \includegraphics[width=0.6\linewidth]{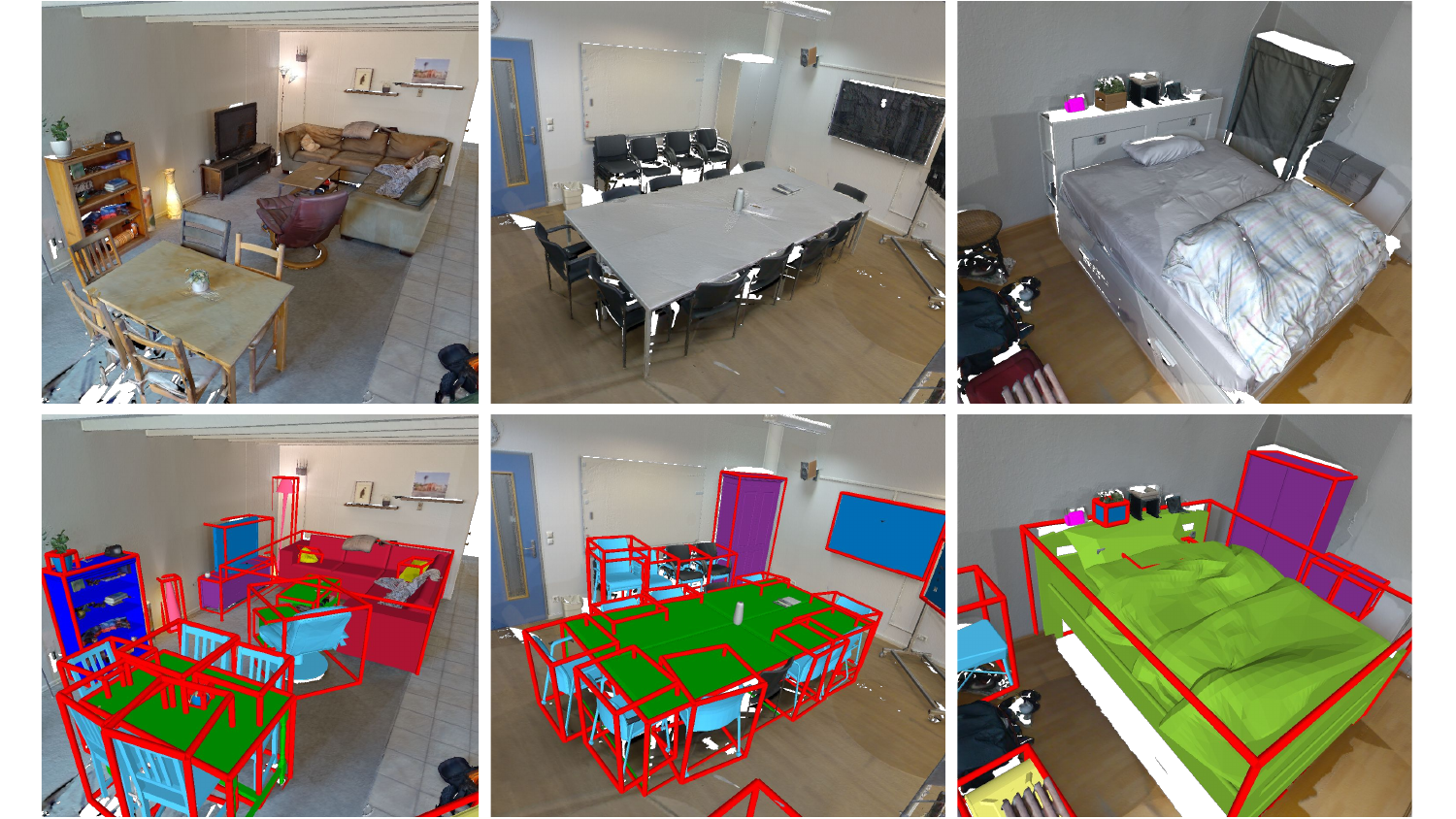} &
        \raisebox{2.5cm}{\begin{tabular}{c}
          \includegraphics[trim={0.05cm 0.05cm 0.0cm 0.2cm},clip,width=0.4\linewidth]{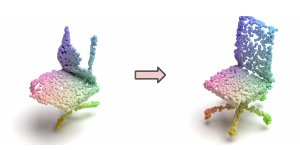} \\[-0.1cm]
          Task 1: Supervised point cloud completion \\
          \includegraphics[trim={0.0cm 0.5cm 0.0cm 0.3cm},clip, width=0.4\linewidth]{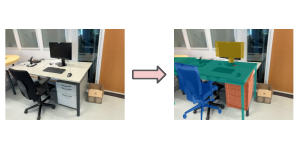} \\[-0.1cm]
          Task 2: Supervised CAD model retrieval and alignment \\
        \end{tabular}}\\[-0.5cm]
        Examples of our annotations for ScanNet++ &\\
    \end{tabular}
    }
    \captionof{figure}{We provide high-quality shape and pose annotations for objects in RGB-D scans for the ScanNet++ dataset. Because our pipeline is automatic, we can annotate almost all objects in the scene even when they are only partially visible, and the annotations are of consistently high quality. 
    We demonstrate that we can use these annotations for supervised learning including point cloud completion~(Task 1) and CAD model retrieval and alignment (Task 2), and that the model performance can be increased either by using the annotations as additional training data, or by using them to fine tune a pre-trained model for a previously unseen dataset.}
\vspace{0.3cm}
    \label{fig:teaser}
}]

\begin{abstract}
\blfootnote{$^*$ The first two authors contributed equally to this work.}

High-level 3D scene understanding is essential in many applications. However, the challenges of generating accurate 3D annotations make development of deep learning models difficult. We turn to recent advancements in automatic retrieval of synthetic CAD models, and show that data generated by such methods can be used as high-quality ground truth for training supervised deep learning models. 
More exactly, we employ a pipeline akin to the one previously used to automatically annotate objects in ScanNet scenes with their 9D poses and CAD models. This time, we apply it to the recent ScanNet++ v1 dataset, which previously lacked such annotations.
Our findings demonstrate that it is not only possible to train deep learning models on these automatically-obtained annotations but that the resulting models outperform those trained on manually annotated data. We validate this on two distinct tasks: point cloud completion and single-view CAD model retrieval and alignment. Our results underscore the potential of automatic 3D annotations to enhance model performance while significantly reducing annotation costs. To support future research in 3D scene understanding, we release our annotations, which we call SCANnotate++, along with our trained models.
\end{abstract}

\section{Introduction}
\label{sec:intro}

3D scene understanding requires precise object recognition, reconstruction, and alignment. A critical building block in such technologies is the generation of high-quality 3D data, enabling use-cases of artificial intelligence applications for perceiving, interpreting, and interacting with virtual environments.

A variety of multi-view datasets have been developed for 3D scene understanding, including SceneNN~\cite{hua2016scenenn}, ScanNet~\cite{dai2017scannet}, ScanNet++~\cite{yeshwanth2023scannetpp}, Matterport3D~\cite{Matterport3D}, and ARKitScenes~\cite{dehghan2021arkitscenes}. However, most of these datasets lack detailed annotations for object shapes, with the exception of ScanNet, for which the Scan2CAD dataset~\cite{avetisyan2019scan2cad} provides 3D alignments of CAD models to target objects. The lack of available annotations can be explained by the challenge of manually annotating 3D shapes which can be difficult and time consuming for the human annotator, and is therefore prone to human inconsistencies and errors. An alternative approach involves creating synthetic datasets~\cite{zheng2020structured3d,roberts2021hypersim}, which is computationally expensive and requires significant resources. Additionally, methods trained in synthetic environments are likely to fail when applied on real~(physical) world scenes, revealing as the well known issue of synthetic-to-real domain gap in deep learning.

In this work, we examine whether automatic approaches for recovering 3D geometry of objects can be used to train deep learning models. In particular, we examine annotations obtained by HOC-Search~\cite{ainetter2024hoc} that jointly performs discrete and continuous search to automatically retrieve CAD models from a large database of synthetic shapes, \eg ShapeNet~\cite{chang2015shapenet}, and aligns them with objects in 3D scenes. The method assumes that a rough estimate of the object's 3D pose is given for the target object, either through manual annotations or using deep neural networks~\cite{Vu_2022_CVPR}, and then performs optimization in image space of the target scene to create an accurate alignment between a retrieved CAD model and the target object. As \cite{ainetter2024hoc} further demonstrates for the ScanNet dataset~\cite{dai2017scannet}, the quality of such annotations is comparable and often better than the manual annotations from the Scan2CAD dataset~\cite{avetisyan2019scan2cad}. This was used by the authors of \cite{avetisyan2019scan2cad} to create the SCANnotate dataset,  which contains CAD model alignments for 18617 objects for 1513 scenes of the ScanNet dataset. 

Here, we first apply HOC-Search on the ScanNet++~v1 dataset~\cite{yeshwanth2023scannetpp} to automatically obtain additional CAD model alignments for 5290 objects from 280 scans. We call this new dataset SCANnotate++. Then, as evaluation of the annotation quality, we show that deep learning approaches can be trained on such automatically annotated data. We consider two tasks that are critical for 3D scene understanding:

First, we tackle point cloud completion and introduce a pipeline with two stages. We train the ShapeGF point cloud auto-encoder network~\cite{cai2020learninggradientfieldsshape} on complete point clouds from the ShapeNet dataset~\cite{chang2015shapenet}. Then, we train a second encoder on real world partial point clouds with the objective to output the same embeddings as the first encoder, hence enabling point cloud completion. 
As our results for the ScanNet dataset~\cite{dai2017scannet} demonstrate, our model trained using automatically generated annotations performs better than the model trained on manually annotated training data from Scan2CAD~\cite{avetisyan2019scan2cad}. 

Second, we consider CAD model retrieval and alignment from single RGB images. We show that ROCA~\cite{gumeli2022roca} trained on the automatically generated SCANnotate dataset performs better than the original model trained on manual annotations from the Scan2CAD dataset, which shows the general usefulness of automatically generated CAD model annotations. Furthermore, we show that it is now possible to train and test ROCA on the ScanNet++ dataset by using our proposed SCANnotate++ dataset, and we are able to present compelling results.  

In summary, this paper provides insights into how pipelines for automatic annotations can be beneficial when training deep learning models for 3D scene understanding. We adapt HOC-Search to automatically obtain high-quality 3D object annotations for indoor scenes. As the result, we introduce SCANnotate++, a new dataset that provides CAD model alignments for objects in the ScanNet++ dataset. In addition, we showcase two successful use-cases for training deep learning methods on such automatically obtained annotations: point cloud completion, and single-view CAD model retrieval and alignment. We will make our dataset and models available, to foster future developments in 3D scene understanding.

\section{Related Work}
\label{sec:related_work}

\begin{table}[]
\centering
\scalebox{0.72}{
\begin{tabular}{@{}ccccc@{}}
\toprule
Dataset   & \begin{tabular}[c]{@{}c@{}} CAD Model\\ Annotations\end{tabular}    &\begin{tabular}[c]{@{}c@{}} \# 3D\\ Scans\end{tabular} & \begin{tabular}[c]{@{}c@{}}   \# Annotated\\ Objects\end{tabular} & \begin{tabular}[c]{@{}c@{}} Annotation\\ Method\end{tabular} \\
\midrule
\midrule
Pix3D~\cite{sun2018pix3d}  & Pix3D~\cite{sun2018pix3d}      & \xmark  & 10k & Manual \\
ARKit Scenes~\cite{dehghan2021arkitscenes} & LASA~\cite{liu2024lasa} & 920 & 11k & Manual \\
ScanNet~\cite{dai2017scannet}  & Scan2CAD~\cite{avetisyan2019scan2cad}     & 1513  & 14k & Manual \\
\midrule
ScanNet~\cite{dai2017scannet}  &  SCANnotate~\cite{ainetter2023automatically,ainetter2024hoc}   & 1513  &  18k & Automatic\\
ScanNet++~v1~\cite{yeshwanth2023scannetpp} & SCANnotate++ (ours) & 280~(*)   & 5k  &Automatic\\ 
\bottomrule
\end{tabular}
}
\caption{\textbf{Comparison of the most popular real RGB-D datasets for indoor scene understanding where CAD model and object pose annotations are available.} (*) At the time of annotating the dataset, only ScanNet++~v1 was publicly available. Hence, the current version of SCANnotate++ provides annotations for all 280 ScanNet++~v1 scans where ground truth 3D segmentation is available. 
}
\label{tab:datasets_overview}
\vspace{-3mm}
\end{table}

\paragraph{CAD model annotations for indoor scans.} 
We compare related popular datasets for indoor 3D scene understanding~\cite{hua2016scenenn,dai2017scannet,Matterport3D,dehghan2021arkitscenes,yeshwanth2023scannetpp,maninis2023cad,liu2024lasa} in Table~\ref{tab:datasets_overview}. For the ScanNet dataset~\cite{dai2017scannet}, Scan2CAD~\cite{avetisyan2019scan2cad} introduces  annotations by manually retrieving CAD models from ShapeNet~\cite{chang2015shapenet} and aligning them to target objects. However, such annotations are very difficult to obtain as they require hard labor from human annotators. Frameworks for generating automatic annotations are becoming an important aspect for tasks in 3D scene understanding~\cite{chibane_inst,Huang2023Segment3D,weder2024labelmaker,ji2024arkit, chen20224dcontrast}. In the context of CAD model representations, CAD-Estate~\cite{maninis2023cad} introduces a semi-automatic framework for RGB video annotation, that first automatically retrieves 10 most promising candidates, which is followed by manual verification, and manual annotation of 2D-to-3D correspondences that are used to align CAD model with the scene. SCANnotate~\cite{ainetter2023automatically,ainetter2024hoc} and HOC-Search~\cite{ainetter2024hoc} go one step further and perform CAD model retrieval and alignment automatically on scenes of the ScanNet dataset. In this work, we examine the aspect of training supervised learning methods on such automatically generated training data. 

\paragraph{Supervised CAD model retrieval.} The availability of CAD model annotations enables a range of applications. For example, the manual Scan2CAD annotations~\cite{avetisyan2019scan2cad} were used to train methods for CAD model alignment from point clouds~\cite{avetisyan2019scan2cad,avetisyan2019end,langer2024fastcad}. These methods assume that a pool of objects in the scene is given and the goal is to align objects from this pool to the objects in the scene. In this work, we evaluate the performance of ROCA~\cite{gumeli2022roca}, a supervised method for CAD model retrieval and alignment from single views. ROCA learns to perform CAD model retrieval from a candidate object pool, and estimates dense correspondences to enable alignment based on differentiable Procrustes optimization.   

\paragraph{Supervised shape completion methods.} CAD model annotations can also be useful for 3D shape completion~\cite{rao2022patchcomplete,wu20153d,dai2017shape,dai2019scan2mesh,li2020self,tang2019skeleton,tang2021skeletonnet}. Supervised shape completion methods learn more general shape representations and use generative modeling formulations to model ambiguities in reconstruction. Some approaches focus on learning representative shape priors from diverse representations, such as volumetric grids~\cite{wu20153d,dai2017shape}, continuous implicit functions~\cite{peng2020convolutional,mescheder2019occupancy,park2019deepsdf, chen2024mesh2nerf}, point clouds~\cite{stutz2020learning, cai2020learninggradientfieldsshape}, and meshes~\cite{dai2019scan2mesh,li2020self,tang2019skeleton,tang2021skeletonnet}, which yield impressive results on the categories used for training. Various learning architectures have been explored. For instance, ShapeGF~\cite{cai2020learninggradientfieldsshape} employs a VQ-VAE backbone to estimate gradient vectors at each point and then integrates these vectors to reconstruct a continuous surface. Luo et al.~\cite{luo2021diffusion} introduce Diffusion Probabilistic Models for point cloud completion to address the challenges of noise and incompleteness in real-world scans, while Kasten et al.~\cite{kasten2023point} leverage a pretrained text-to-image diffusion model to complete the surface representation based on a test prompt.

\textvars{dpt,Sil,CD,MSCD}

\begin{figure*}[ht!] 
      \centering
 		\includegraphics[trim={0.1cm 1cm 0cm 0.cm},width=0.9\linewidth]{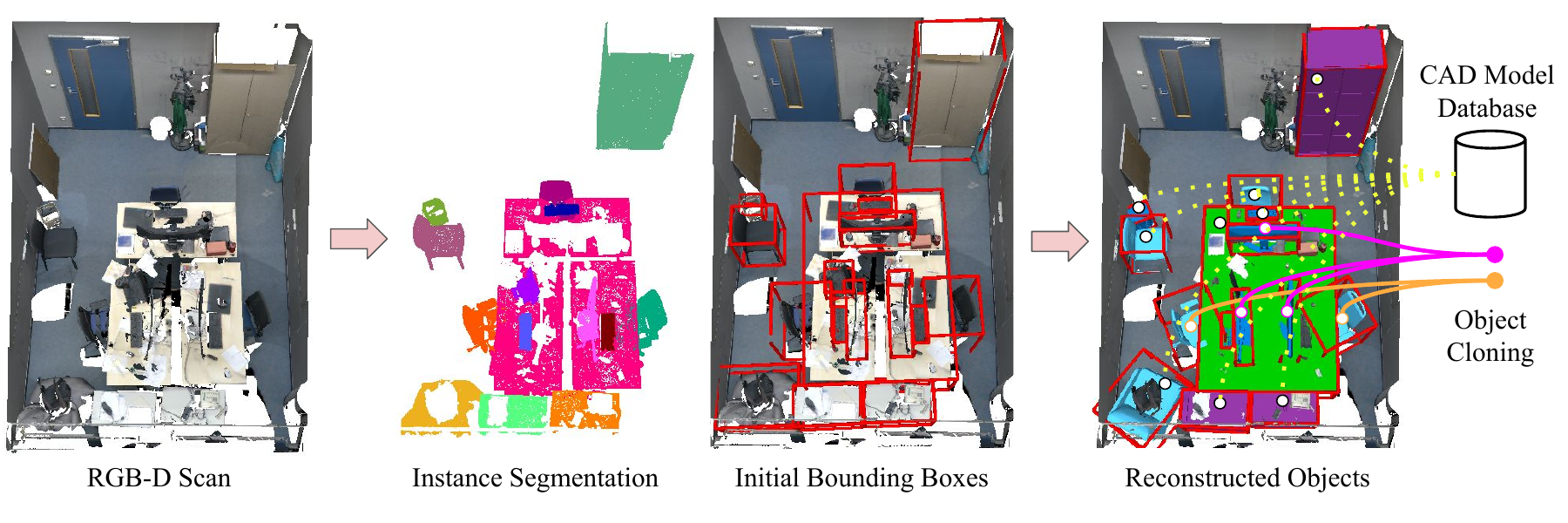}
	\caption[]{\textbf{Our automatic annotation pipeline.} For a given RGB-D scan, we first use the provided 3D object instance segmentation to estimate the pose of each object, visualized as red bounding boxes. The scan, 3D segmentation and initial estimated poses are then used as input for our annotation method. As final result, we provide high quality 3D shape annotations in the form of CAD models retrieved from a large shape database, and corresponding 9D pose annotations for all target objects. }
	\label{fig:overview}
    \vspace{-3mm}
\end{figure*}

\section{Automatic Generation of CAD Annotations}
\label{sec:method}

Given an RGB-D scan of a static scene, our goal is to generate 3D shape and 9D pose annotations for the objects in the scene. Based on findings from \cite{ainetter2023automatically,ainetter2024hoc}, we propose an automatic annotation pipeline that relies on given 3D semantic object instance segmentation to identify the target objects to be annotated, and then utilizes CAD models from a large database for shape retrieval.

To be precise, we use the annotation pipeline proposed in~\cite{ainetter2023automatically}, but replace the original shape retrieval algorithm with the more effective HOC-Search~\cite{ainetter2024hoc} algorithm. For HOC-Search, our only modification compared to the original work is the adaptation of the threshold for CAD model selection and pose refinement, which led to more accurate annotations in our experiments.

Figure~\ref{fig:overview} gives an overview of our annotation process.
We achieve high-quality results, in terms of 3D fitness with respect to the target object, and in terms of alignment of re-projections with the corresponding RGB-D frames of the target scan.

In the following sections, we discuss the main parts of our pipeline, including data pre-processing for initial object pose estimation, our shape and pose retrieval algorithm and the refinement of the retrieval results. Then, we describe our manual verification process that ensures the quality of the final annotations. Finally, we provide details and statistics of our resulting SCANnotate++ dataset.

\subsection{CAD Model-Retrieval and Pose Reconstruction}

Our reconstruction pipeline, based on SCANnotate~\cite{ainetter2023automatically,ainetter2024hoc}, consists of two main parts: In the first part, we perform CAD model retrieval. Unlike SCANnotate which relies on exhaustive CAD model search, we utilize HOC-Search~\cite{ainetter2024hoc} as our retrieval method. In the second part, we apply CAD model clustering and cloning and pose refinement for cloned CAD models. Afterwards, we perform manual inspection of the annotation results and, if needed, perform manual re-annotation of outliers to ensure high quality results of our annotation. 

\subsubsection{Preprocessing: Object Pose Initialization}
\label{sec:box_preprocessing}

Given the 3D instance segmentation of a target object and the corresponding class label, we  first calculate a 3D bounding box as initialization for position and scale of the object. For each object, we estimate an oriented bounding box using Trimesh~\cite{trimesh}, whereas the bounding box is aligned with the major axes of the target object, and is utilized for pose initialization. The initial pose does not have to be very precise, as the CAD model pose is later optimized during the CAD model retrieval process.

\subsubsection{Initial CAD Model and Pose Annotation}

After obtaining the initial pose of the target object, we retrieve a CAD model from a shape database and predict its corresponding 3D pose to align well to the observations of the target object. We use HOC-Search~\cite{ainetter2024hoc} as retrieval method, which retrieves good results even in challenging scenarios and in presence of noise, \eg when objects are only partially visible, or when depth sensor values are inaccurate or missing. 

HOC-Search performs joint optimization of discrete and continuous parameters, a feature that is crucial to solve our annotation task. Its algorithm can be used in combination with an objective term solely dependent on data observations (for example render-and-compare or chamfer distance), and does not rely on learning. It generalizes very well and is well suited for data annotation, in contrast to 3D deep learning approaches that are prone to errors when performing inference on out-of-distribution data. In this section, we provide an overview, and we refer to the original HOC-Search~\cite{ainetter2024hoc} paper for a more detailed description about the algorithm and the implementation.

HOC-Search is based on two main components, a data structure called HOC-Tree and an efficient algorithm to search this structure. HOC-Tree represents discrete parameters as nodes of a tree. \cite{ainetter2024hoc} proposes to organize a set of CAD models into a tree structure based on hierarchical clustering of shape similarities. Therefore,  the similarity of shapes increases with the depth of the tree, which makes HOC-Tree a very useful data structure in practice. Then, HOC-Search adapts a variant of Monte Carlo Tree Search (MCTS)~\cite{stekovic2021montefloor,MonteRoom} that efficiently searches tree structures by jointly optimizing discrete and continuous parameters in an iterative search.

\paragraph{Objective term for optimization.}
One iteration of HOC-Search returns a CAD model and a 3D pose, which is directly inferred from the nodes of the corresponding path through the tree. To evaluate the quality of this solution, we calculate an objective term based on render-and-compare that was originally proposed in~\cite{ainetter2023automatically}. The render-and-compare objective term $\calL_{RC}$ can be described as 
\begin{equation}
\calL_\text{RC} = \lambda_\dpt \calL_\dpt + \lambda_\Sil \calL_\Sil + \lambda_\CD \calL_\CD \> ,
\label{eq:rac}
\end{equation}
where $\calL_\dpt$ is the L1-distance between depth maps of the target object ant the CAD model, $\calL_\Sil$ is the IoU between silhouettes of target object and CAD model, and $\calL_\CD$ defines the single-direction chamfer distance, with $\lambda_\dpt$, $\lambda_\Sil$ and $\lambda_\CD$ the corresponding weights. The weights and the hyperparameters for MCTS have been taken from~\cite{ainetter2024hoc}. For each target object we use a maximum number of 1200 MCTS iterations, which empirically showed to be sufficient to converge to a good solution.

\paragraph{Search algorithm: Adapted threshold for CAD model selection and pose refinement.}
Because the initial pose parameters which are encoded as nodes in the HOC-Tree are discretized, and due to the fact that the shape retrieval relies on a correct initial pose, it would limit the performance of HOC-Search to directly use the object pose based on the discrete parameters. Therefore, HOC-Search implements a pose refinement step, which uses gradient-based optimization to iteratively adapt the initial pose for the selected CAD model after every HOC-Search iteration. As proposed in the HOC-Search paper, we use the render-and-compare objective term as defined in Equation~\ref{eq:rac} to optimize the 9-DOF pose parameters, using the differentiable rendering pipeline of \cite{ravi2020pytorch3d} and the Adam optimizer~\cite{kingma2014adam}. 

Note that HOC-Search performs this pose refinement step after every HOC-Search iteration for which the resulting CAD model provides a better score according to the objective term. We modify this criterion: We empirically found it beneficial to additionally consider CAD models for pose refinement which lead to a slightly worse score (if the new score is smaller than $1.1$ times the previous best score). While this modification effectively increases the computation time as more candidates are selected for pose refinement, these additional solutions led in some cases to the most suitable result after pose refinement hence increasing the overall quality of our retrieved annotations.

\begin{figure} 
      \centering
 		\includegraphics[trim={4cm 12cm 12cm 2cm},width=1\linewidth]{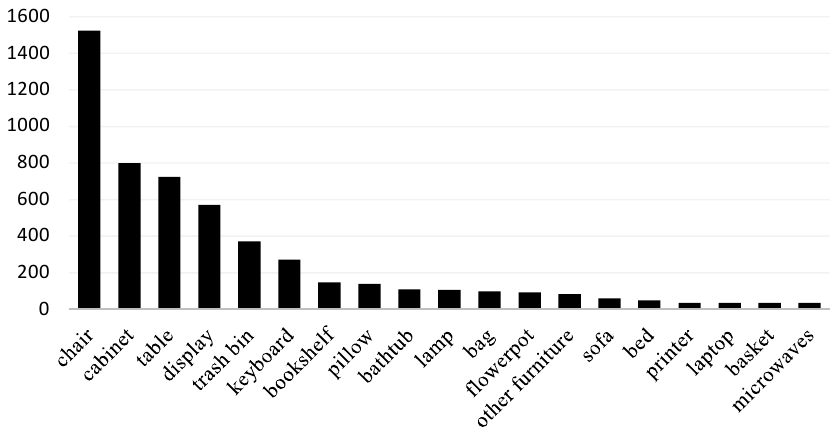}
	\caption[]{\textbf{Class histogram of the SCANnotate++ dataset.} The histogram shows the typical long tail distribution of objects in indoor scenes, whereas common classes like chair, cabinet, and table make up the majority of the annotated objects.}
	\label{fig:data_statistics}
    \vspace{-3mm}
\end{figure}

\subsubsection{Final Shape and Pose Refinement}

 After per-object CAD model and pose annotation for all objects in a scene, we exploit the prior knowledge that multiple objects of the same class in one scene are likely to have the same shape. We use CAD model clustering and cloning proposed in \cite{ainetter2023automatically} for the classes chair, cabinet, sofa, bookshelf, display and table. The aim is to cluster all CAD models which belong to the same class according to their shape similarity, using the chamfer distance to quantify shape similarity. Afterwards, all target objects which belong to the same cluster are replaced with the one common CAD model which minimizes the sum of the objective term in Equation~\ref{eq:rac}. As last step, we perform the pose refinement step---the same as HOC-Search---for all CAD models which have been effected by the CAD model cloning, to ensure that the final pose is optimal for the CAD model which represents the shape of the target object.

\subsubsection{Manual Quality Verification}

We visually check the quality of shape and pose for all objects after automatic annotation, and manually edited them when needed. For chairs, the most complex class in terms of geometry,  we had to manually adapt 68 of the 1523 annotated objects, which is roughly 4.5\%.

There are indeed some scenarios where our pipeline can produce annotations of insufficient accuracy: If the initial 3D object segmentation or the extracted bounding box proposals are significantly wrong, our pipeline is not able to produce suitable annotations. Furthermore, as our objective term is based on captured sensor data, the quality of the retrieved annotations is directly related to quality of the scan: severe occlusions and partial object captures in input scenes affect the annotation quality. 

As part of the manual quality verification, if the rotation around the up-axis of a specific annotation is off by either $[0^\circ,90^\circ,180^\circ,270^\circ]$ we manually adapted it. This failure case can happen for box-like object shapes (for example for classes like cabinet, stove, and washer, because they often look almost similar from each side). For cases where the shape of the retrieved object does not match the shape of the target object well, we manually re-annotated the target object with a suitable CAD model.
In either of the aforementioned cases where we manually adapted the pose or shape of the annotation, we afterwards re-run the object pose refinement step to align the annotation to the data observations.

When the initial ground truth instance segmentation from the dataset was not of sufficient quality, we remove our corresponding annotation.

\begin{figure}[t!]
    \centering
    \scalebox{0.95}{
    \begin{tabular}{ccc}
        \includegraphics[width=0.31\linewidth]{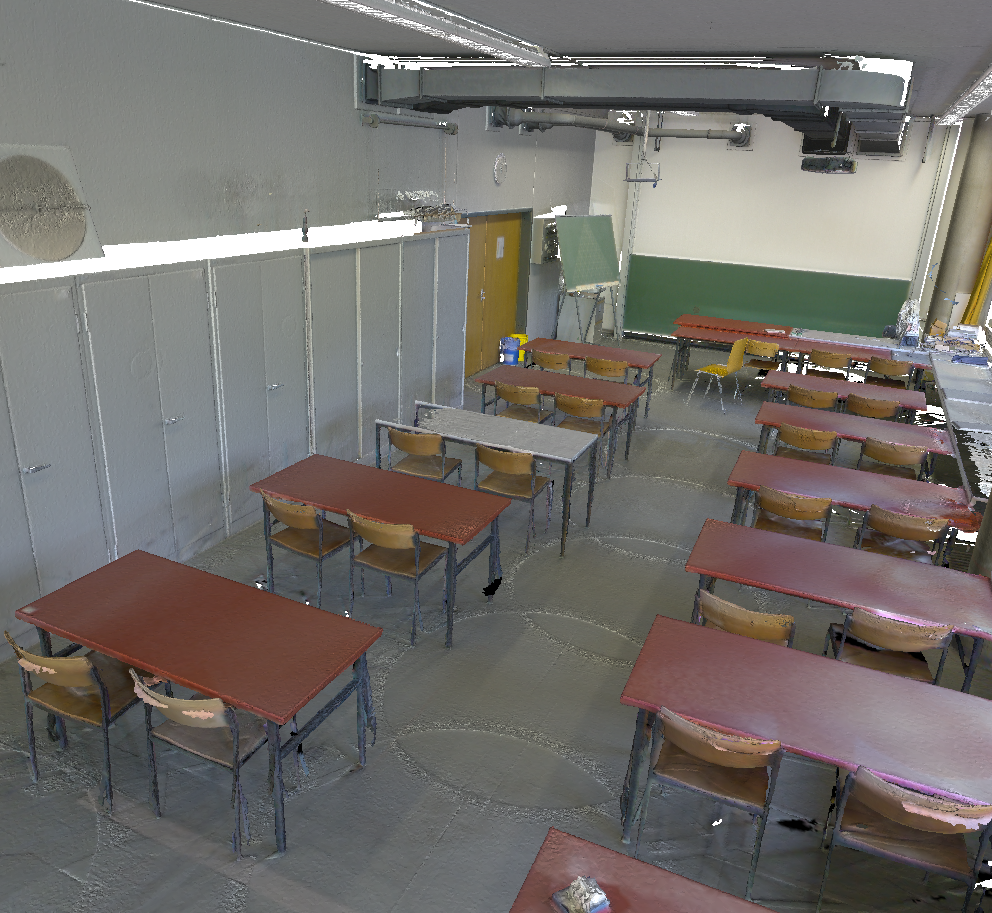} &
        \includegraphics[width=0.31\linewidth]{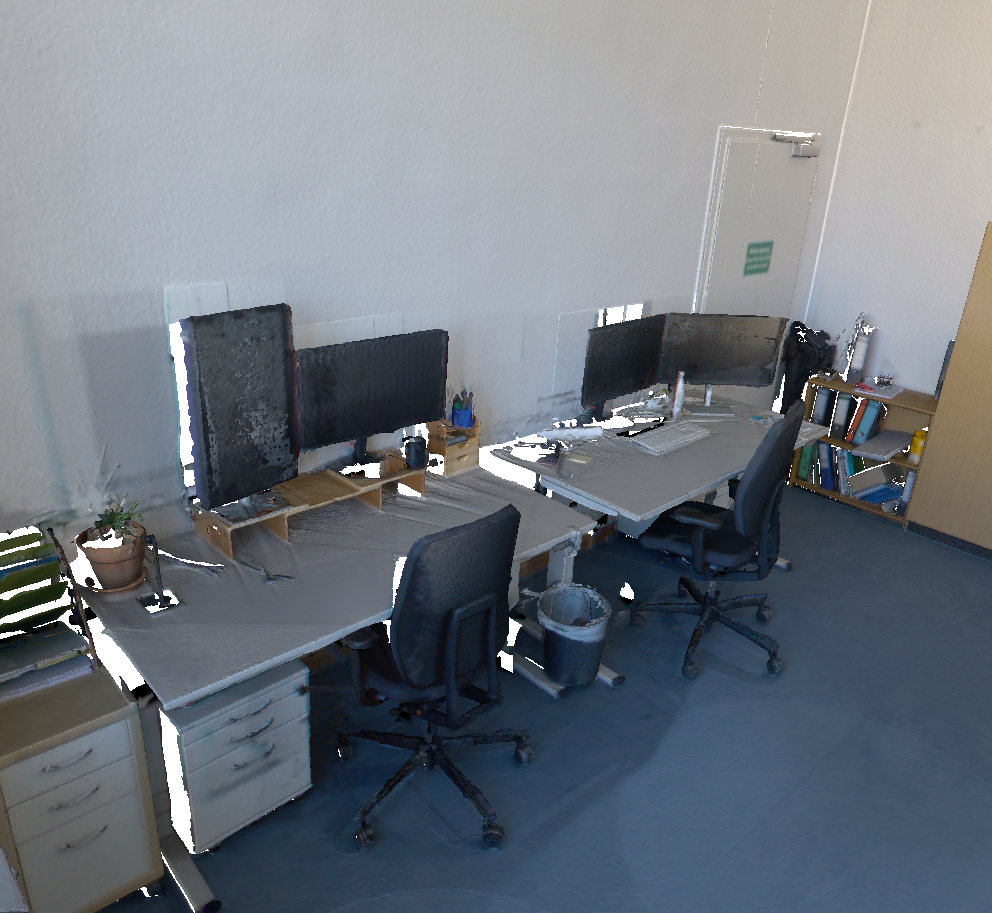} &
        \includegraphics[width=0.31\linewidth]{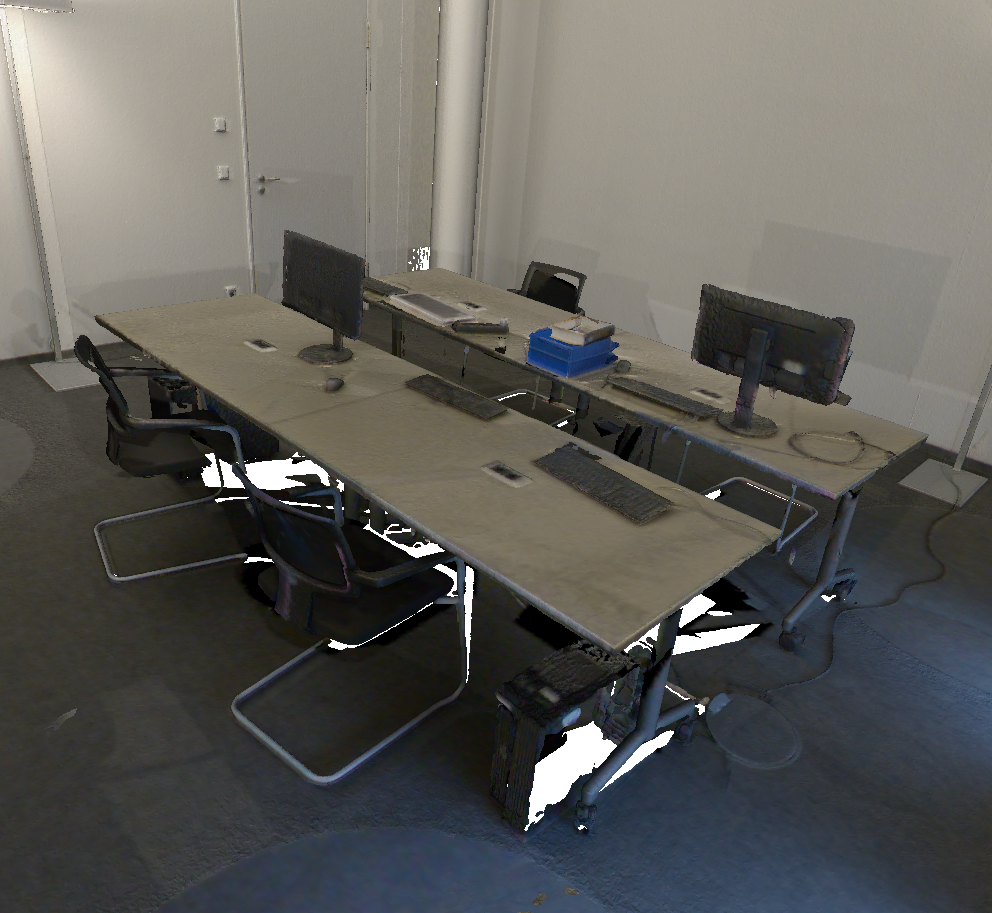} \\
        \multicolumn{3}{c}{(a) ScanNet++~\cite{yeshwanth2023scannetpp} scans (3D meshes)} \\
        \includegraphics[width=0.31\linewidth]{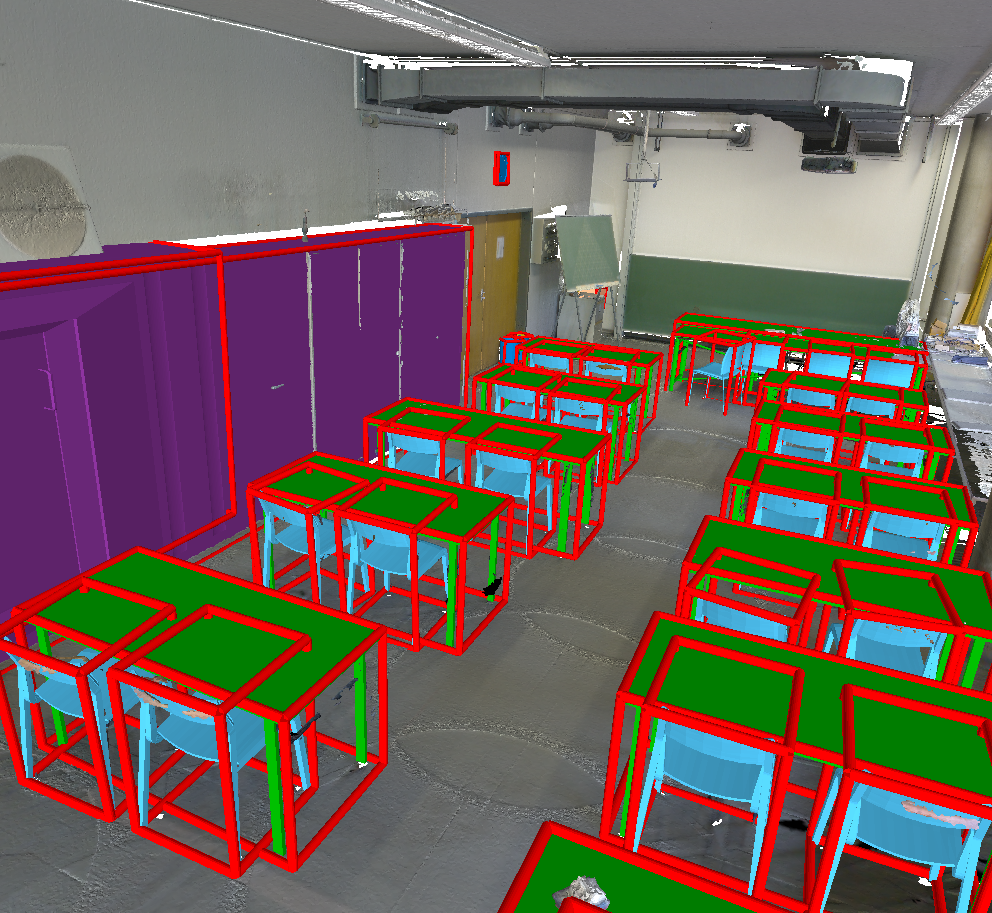} &
        \includegraphics[width=0.31\linewidth]{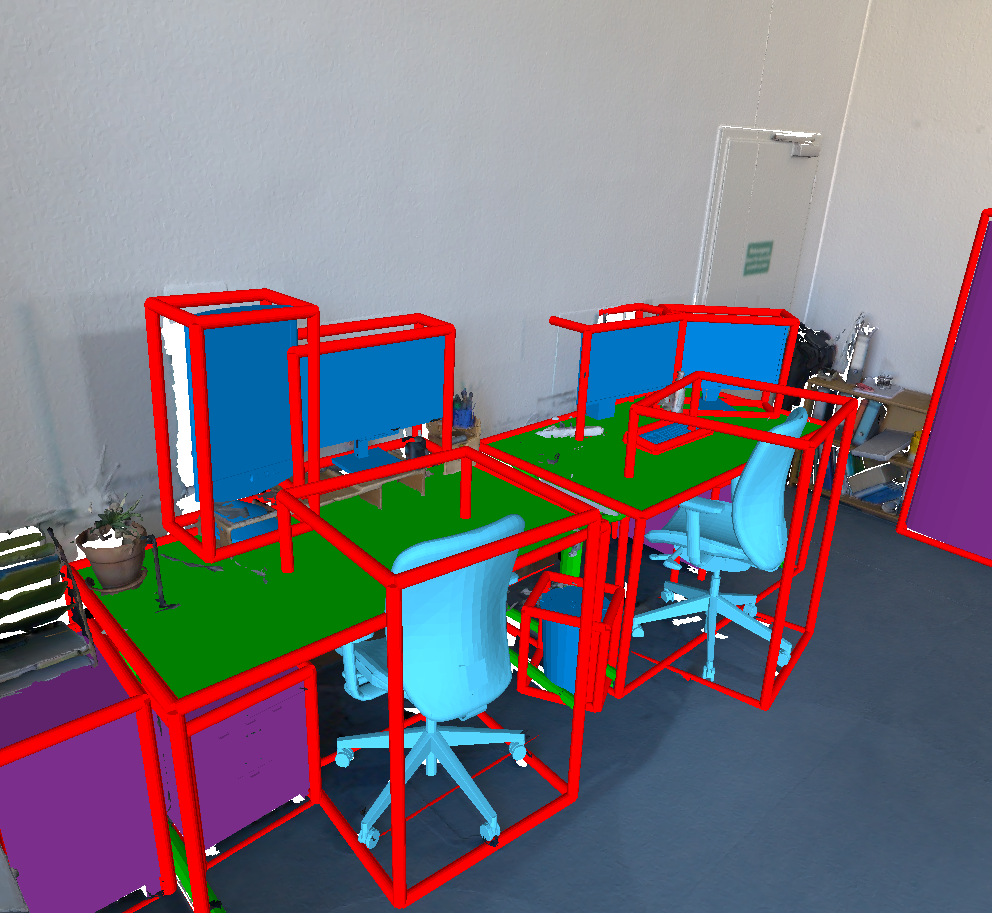} &
        \includegraphics[width=0.31\linewidth]{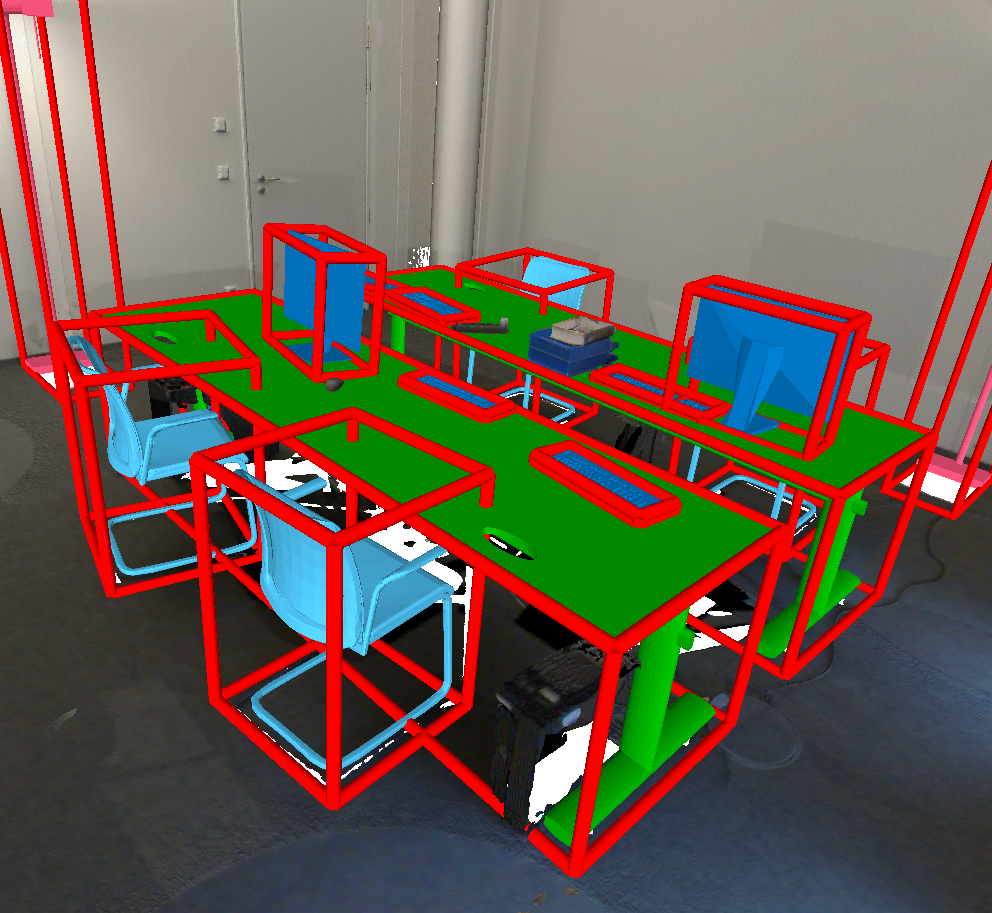} \\
        \multicolumn{3}{c}{(b) Our automatic CAD and pose annotations} \\
        \includegraphics[width=0.31\linewidth]{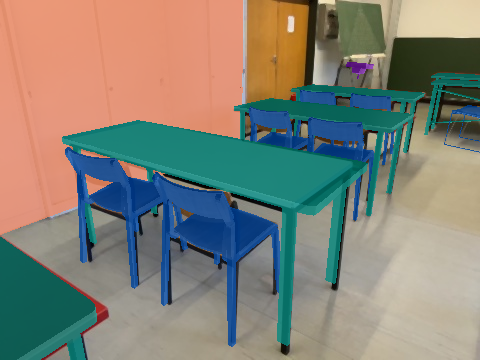} &
        \includegraphics[width=0.31\linewidth]{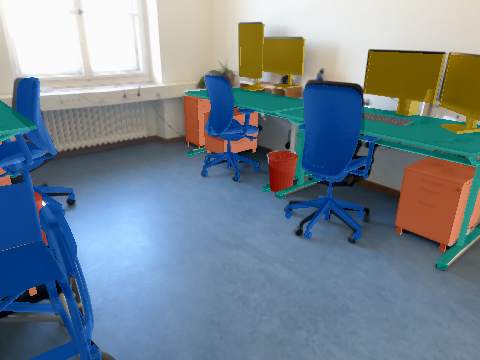} &
        \includegraphics[width=0.31\linewidth]{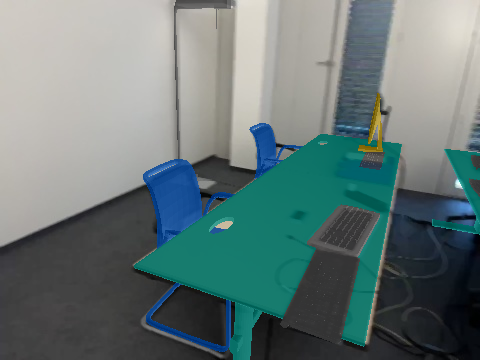} \\
        \multicolumn{3}{c}{(c) Our annotations projected into sampled scene images} \\
        \end{tabular}
        }
        \vspace{-2mm}
    \caption{\textbf{Examples of our SCANnotate++ annotations.} Our annotations accurately capture the 3D geometry of target objects, and their re-projections in the images align accurately with the objects.} 
    \label{fig:sample}
    \vspace{-5mm}
\end{figure}

\subsection{The SCANnotate++ Dataset}

The resulting \textbf{SCANnotate++} dataset was generated using our annotation pipeline to annotate 280 scans from the popular ScanNet++~v1 dataset~\cite{yeshwanth2023scannetpp} providing 5290 CAD model and pose annotations. Figure~\ref{fig:sample} highlights the quality of our SCANnotate++ annotations. We rely on ground truth object instance segmentations from ScanNet++, and use ShapeNet~\cite{chang2015shapenet} as our CAD model database which contains more than 32k objects for the classes in ScanNet++. 

Figure~\ref{fig:data_statistics} shows the distribution of annotated objects over the classes. In total, we annotated objects for 26 different semantic classes where CAD models have been available in the ShapeNet~\cite{chang2015shapenet} database. 

Because our CAD retrieval process is automatic and the objective term is based on observed data, SCANnotate++ offers a high diversity of 3D shapes. For example for the class chair, our pipeline selected 376 chair instances for the 1523 annotated objects, with a total of 6779 chair models available in the ShapeNet database.

\section{Experiments}

In this section, we evaluate the performance of supervised learning methods trained on automatically generated CAD model annotations for ScanNet~\cite{dai2017scannet} and ScanNet++~\cite{yeshwanth2023scannetpp}. More specifically, we focus on tasks of point cloud completion and single-view CAD model retrieval and alignment.

\paragraph{Datasets.} 
We use the automatic annotations from SCANnotate~\cite{ainetter2023automatically,ainetter2024hoc} for the ScanNet dataset~\cite{dai2017scannet} and our SCANnotate++ annotations for the  ScanNet++~v1 dataset~\cite{yeshwanth2023scannetpp} as two complementary resources of high quality for training supervised learning methods. To demonstrate the feasibility and effectiveness of automatic annotations, we compare supervised learning methods trained on automatically generated data against manual annotations from Scan2CAD~\cite{avetisyan2019scan2cad}, which remains a widely adopted benchmark for annotations on the ScanNet dataset~\cite{dai2017scannet}. 

Table~\ref{tab:data} provides a detailed breakdown of the number of training and testing objects across all categories for each dataset. For Scan2CAD and SCANnotate, we utilize the training and testing scene splits defined by the ScanNet framework. Similarly, for SCANnotate++, we adhere to the scene split settings specified in ScanNet++~v1.

\begin{table}[]
\centering
\scalebox{.9}{
\begin{tabular}{@{}cccc@{}}
\toprule
Dataset & \multicolumn{2}{c}{Number of objects} \\
\cmidrule{2-3}
  & Training  & Testing \\
\midrule
Scan2CAD~\cite{avetisyan2019scan2cad}  & 10893 & 3109 \\
SCANnotate~\cite{ainetter2023automatically,ainetter2024hoc}  & 14424 & 4193 \\
SCANnotate++ (ours) & 4343 & 947 \\
\bottomrule
\end{tabular}
}
\vspace{-3mm}
\caption{The number of annotated objects used for training and testing for the Scan2CAD~\cite{avetisyan2019scan2cad}, SCANnotate~\cite{ainetter2023automatically,ainetter2024hoc} and SCANnotate++ (ours) datasets.}
\label{tab:data}
\vspace{-3mm}
\end{table}

\subsection{Point Cloud Completion} 

\begin{figure}
\centering
\includegraphics[width=\linewidth]{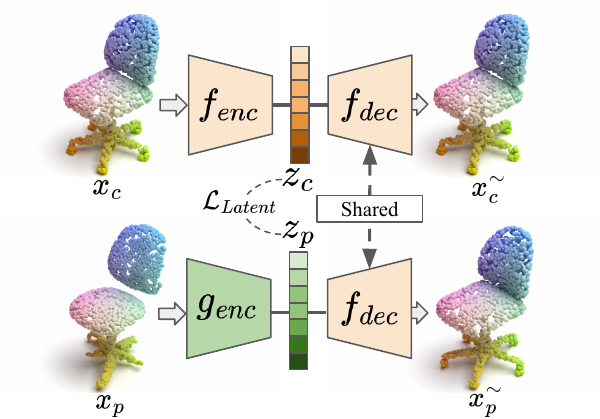}
\vspace{-5mm}
\caption{\textbf{Overview of our point cloud completion pipeline.} First, we train an auto-encoder based on ShapeGF~\cite{cai2020learninggradientfieldsshape} using complete point clouds $x_{c}$. Next, we introduce a separate encoder $g_{enc}$ for partial input point clouds $x_{p}$ from ScanNet~\cite{dai2017scannet} and ScanNet++~\cite{yeshwanth2023scannetpp} and we train it to minimize the difference between latent representations of the ground truth point cloud $z_{c}$ and partial point cloud $z_{p}$. The decoder $f_{dec}$ from the pretrained auto-encoder reconstructs the complete point cloud $x_{\tilde{p}}$ from latent $z_{p}$.} 
\label{fig:method}
\vspace{-3mm}
\end{figure}

Given an input partial point cloud of objects in ScanNet and ScanNet++, we want to reconstruct the missing 3D points. We use automatic annotations from SCANnotate and SCANnotate++ to generate complete point clouds as ground truth for supervised learning of point cloud completion and present our architecture in Figure~\ref{fig:method}.

\paragraph{Our model.}We train our model in two phases. In the first phase, we train an auto-encoder following the ShapeGF framework~\cite{cai2020learninggradientfieldsshape} for point cloud completion, using ground truth point clouds sampled from annotated meshes. Here, encoder $z_c = f_\text{enc}(x_{c})$ encodes a complete point cloud $x_{c}$ into a latent $z_c$, and decoder $x^{\sim}_{c} = f_{dec}(z_c)$ decodes point cloud $x^{\sim}_{c}$ from $z_c$. In the second phase, we introduce a second encoder $z_p = g_\text{enc}(x_p)$, with the same architecture as $f_\text{enc}(\cdot)$, to encode partial input point clouds $x_p$ into latent $z_p$. We use a weighted combination of Mean Squared Error~(MSE) and Kullback-Leibler (KL) divergence: 
\begin{equation}
\calL_\text{Latent} = \lambda_\text{MSE} \calL_\text{MSE}(z_p, z_c) + \lambda_\text{KL} \calL_\text{KL}(z_p,z_c) \> ,
\label{eq:loss}
\end{equation}
where $\lambda_\text{MSE}$ is set to 1 and $\lambda_\text{KL}$ is set to 0.5. Hence, we enforce $z_p$ to be consistent with $z_c$. At inference, the shared fixed decoder $f_\text{dec}(z_p)$ from the pretrained auto-encoder is used to reconstruct the complete point cloud $x^{\sim}_{p}$.

\begin{table}
    \centering
    \scalebox{0.75}{
    \begin{tabular}{cccc}
     \toprule
     Test set & Train set & CD ($\times10^4$) $\downarrow$ & EMD ($\times10^2$) $\downarrow$ \\
     \hline
     Scan2CAD~\cite{avetisyan2019scan2cad} & Scan2CAD~\cite{avetisyan2019scan2cad} & 54.55 & 13.94 \\
     Scan2CAD~\cite{avetisyan2019scan2cad} & SCANnotate~\cite{ainetter2023automatically,ainetter2024hoc} & \textbf{38.16} & \textbf{9.84} \\
     \hline
     SCANnotate & SCANnotate & 42.66 & 10.23 \\
     SCANnotate & SCANnotate,  & \textbf{38.52} & \textbf{9.79} \\
                & SCANnotate++ \\
    \hline
    SCANnotate++ & SCANnotate,  & 40.18 & 9.92 \\
    SCANnotate++ & SCANnotate \&  & \textbf{37.43} & \textbf{9.65} \\
                & SCANnotate++ \\
     \bottomrule
    \end{tabular}}
    \caption{\textbf{Point cloud completion results on scenes of ScanNet~\cite{dai2017scannet} and ScanNet++~\cite{yeshwanth2023scannetpp}.} Training our network on automatic annotations from SCANnotate~\cite{ainetter2023automatically,ainetter2024hoc} results in much better performance compared to training on manually annotated Scan2CAD~\cite{avetisyan2019scan2cad}. Training on both SCANnotate and our SCANnotate++ further improves performance of our network.}
    \label{tab:completion}
\end{table}

\begin{figure*}[ht!] 
    \centering
    \scalebox{.8}{
    \begin{tabular}{cccccc}

         \includegraphics[trim={15cm 0.0cm 15cm 0.0cm},clip,width=0.15\textwidth]{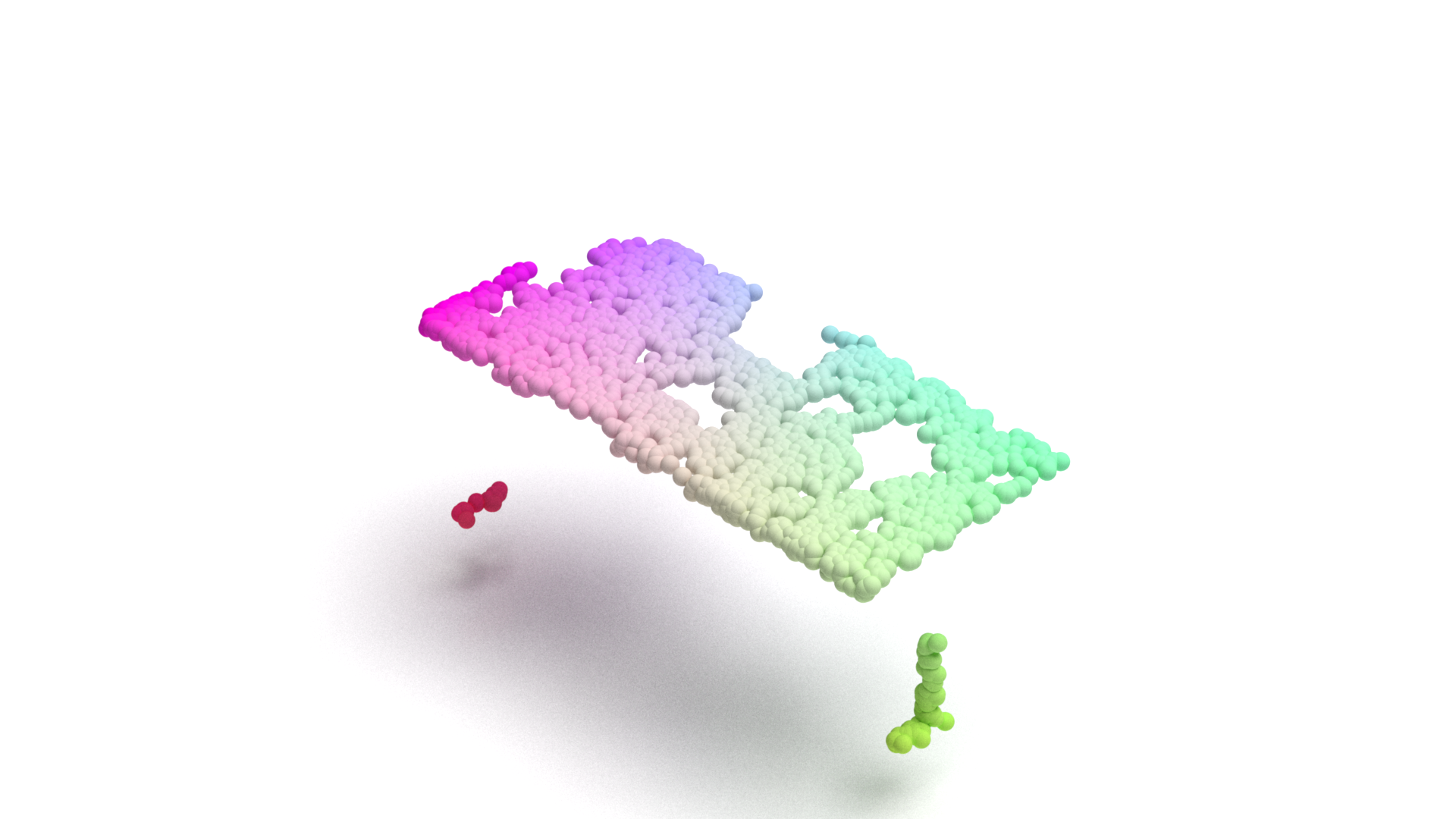} &
        \includegraphics[trim={15cm 0.0cm 15cm 0.0cm},clip, width=0.15\textwidth]{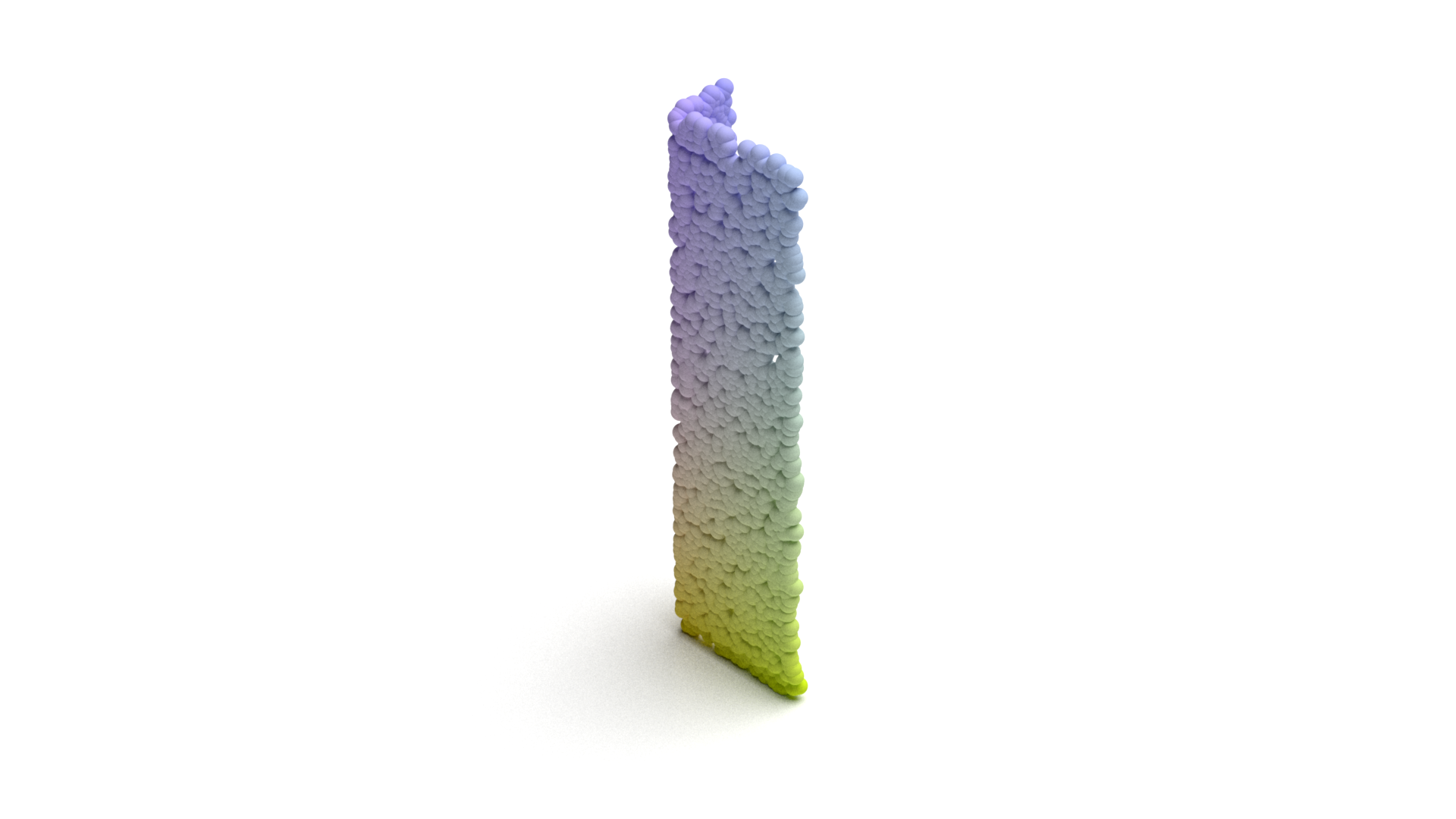} &
        \includegraphics[trim={15cm 0.0cm 15cm 0.0cm},clip, width=0.15\textwidth]{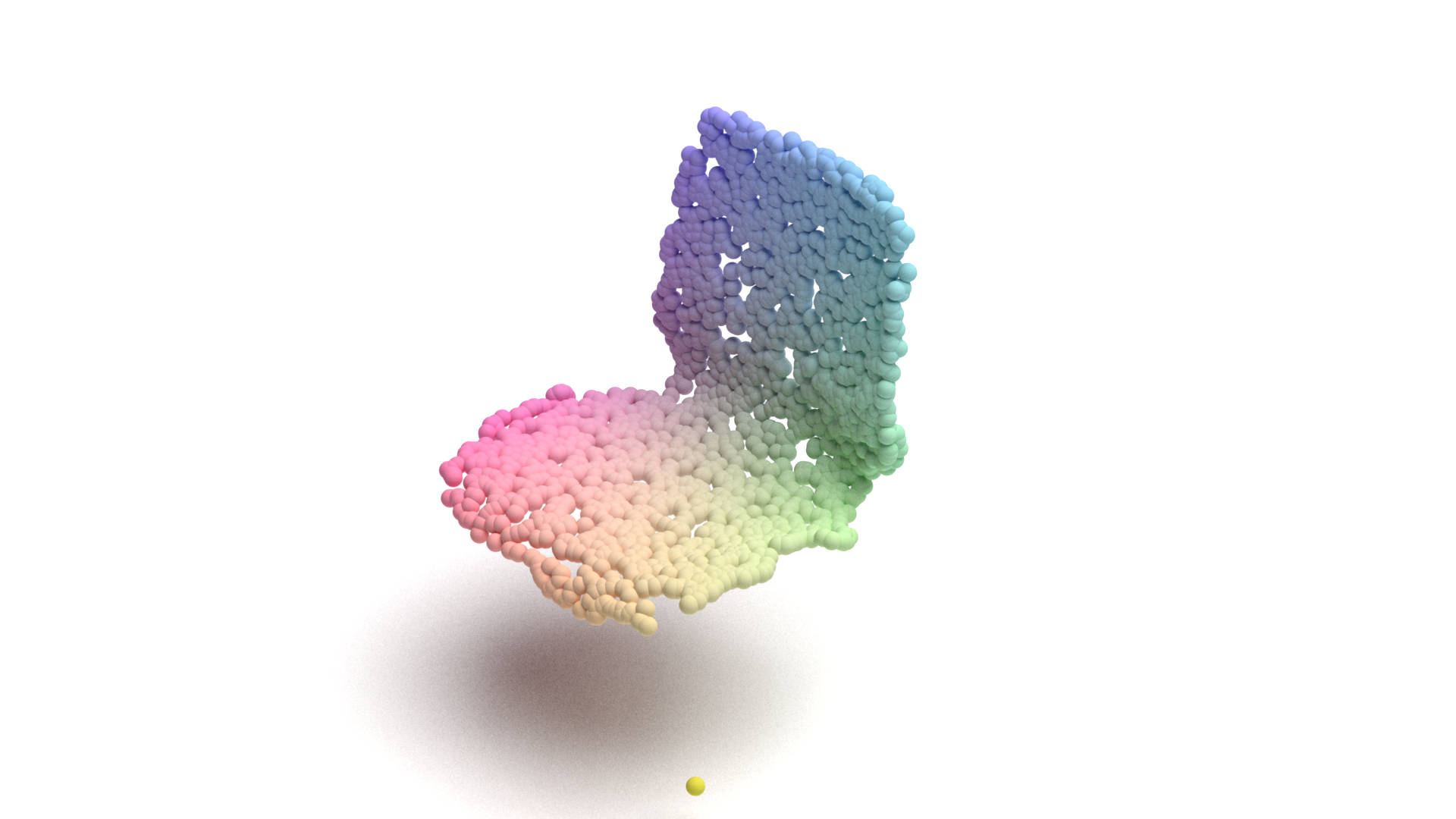} &
        \includegraphics[trim={15cm 0.0cm 15cm 0.0cm},clip, width=0.15\textwidth]{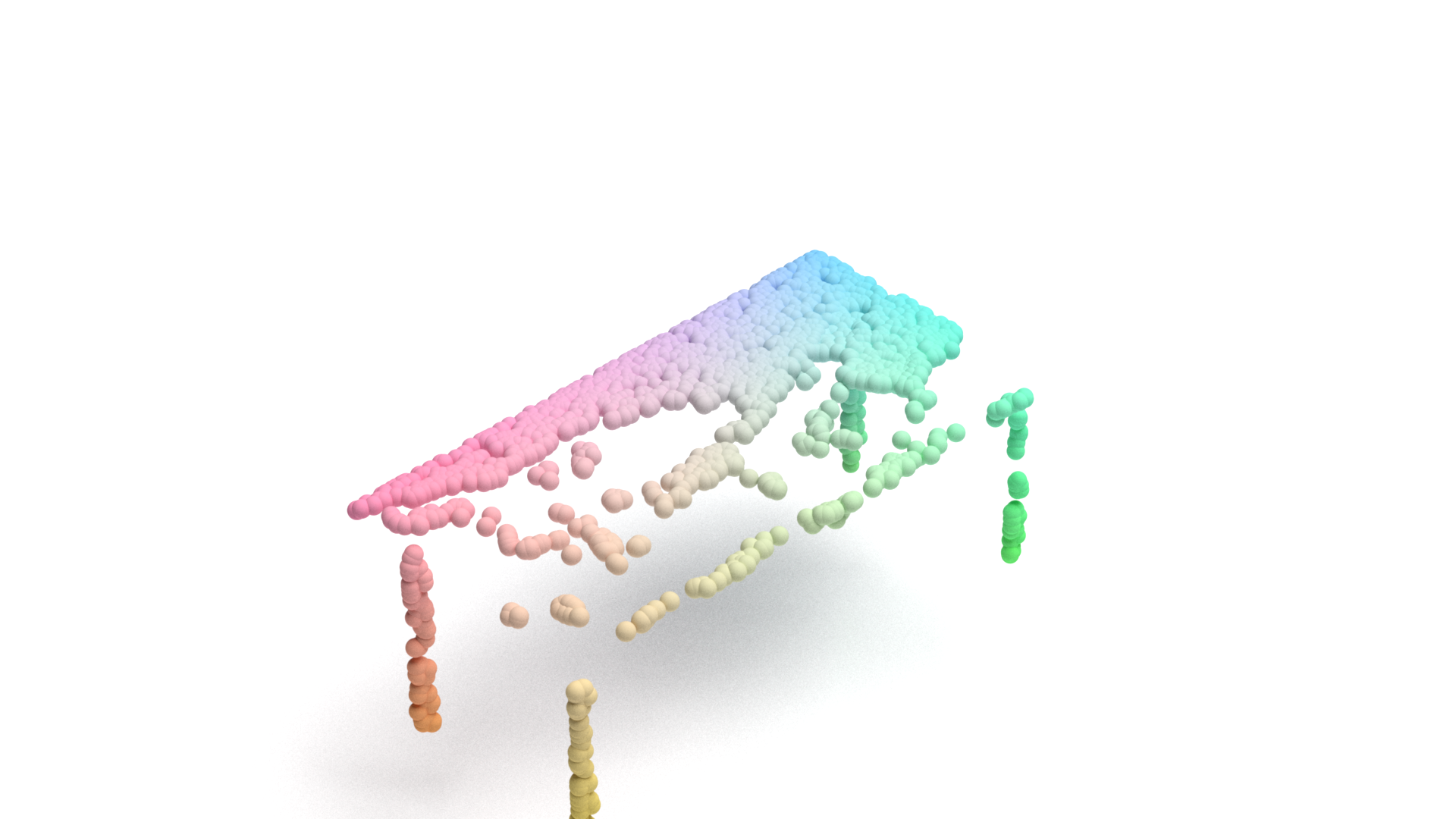} &
        \includegraphics[trim={15cm 0.0cm 15cm 0.0cm},clip,width=0.15\textwidth]{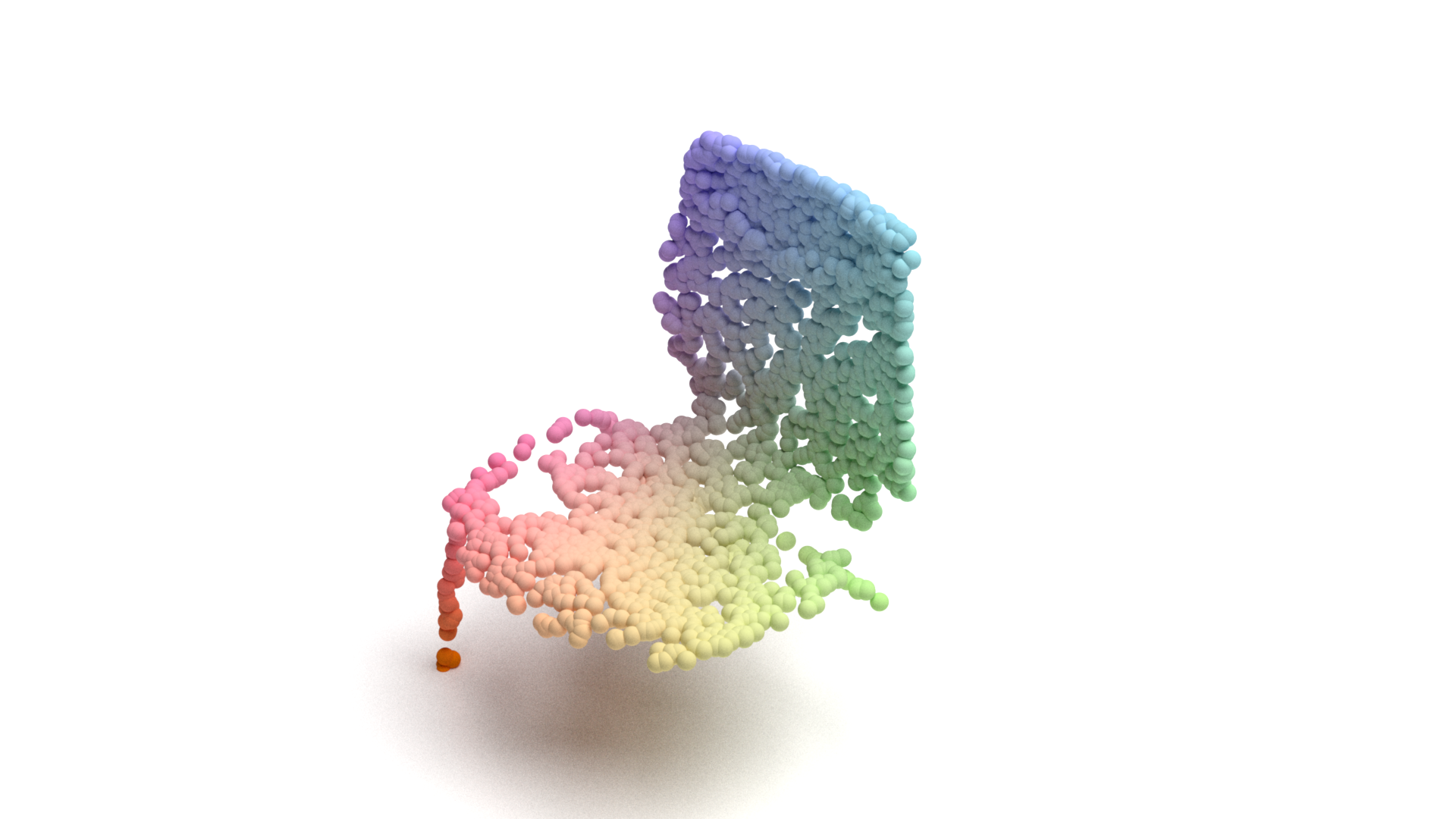} &
        \includegraphics[trim={15cm 0.0cm 15cm 0.0cm},clip,width=0.15\textwidth]{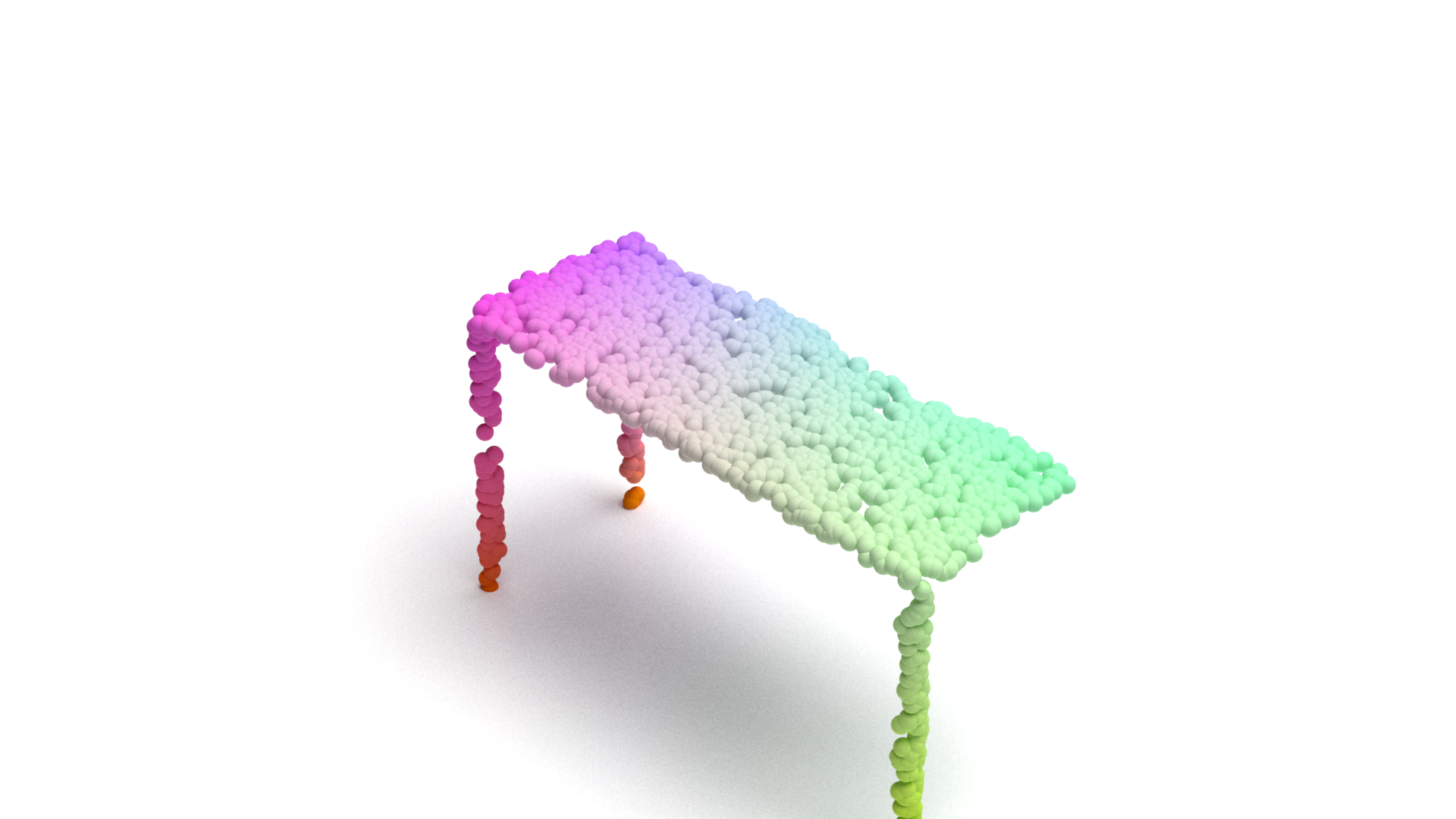} \\
        \includegraphics[trim={15cm 0.0cm 15cm 0.0cm},clip, width=0.15\textwidth]{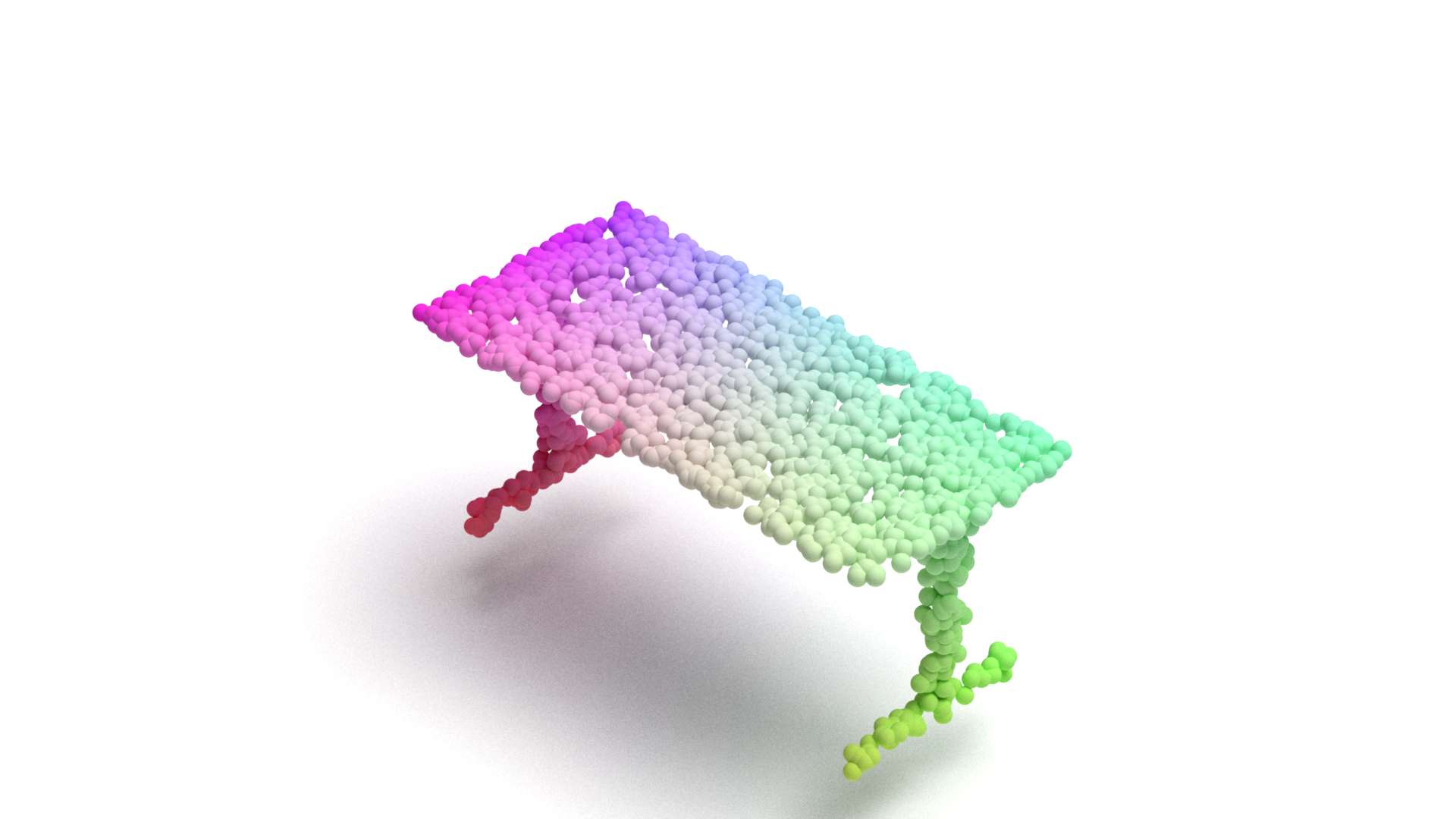} &
        \includegraphics[trim={15cm 0.0cm 15cm 0.0cm},clip, width=0.15\textwidth]{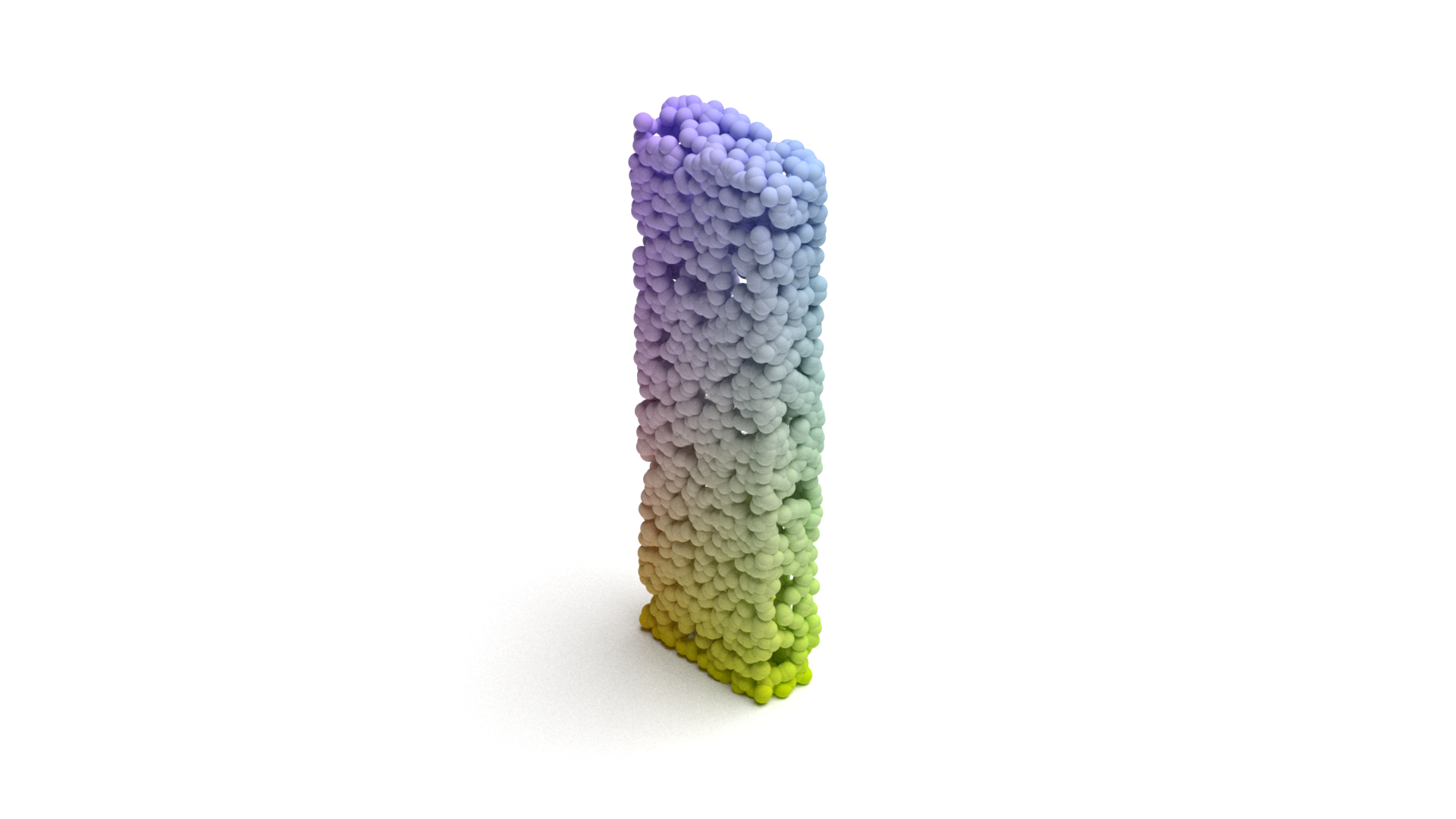} &
        \includegraphics[trim={15cm 0.0cm 15cm 0.0cm},clip, width=0.15\textwidth]{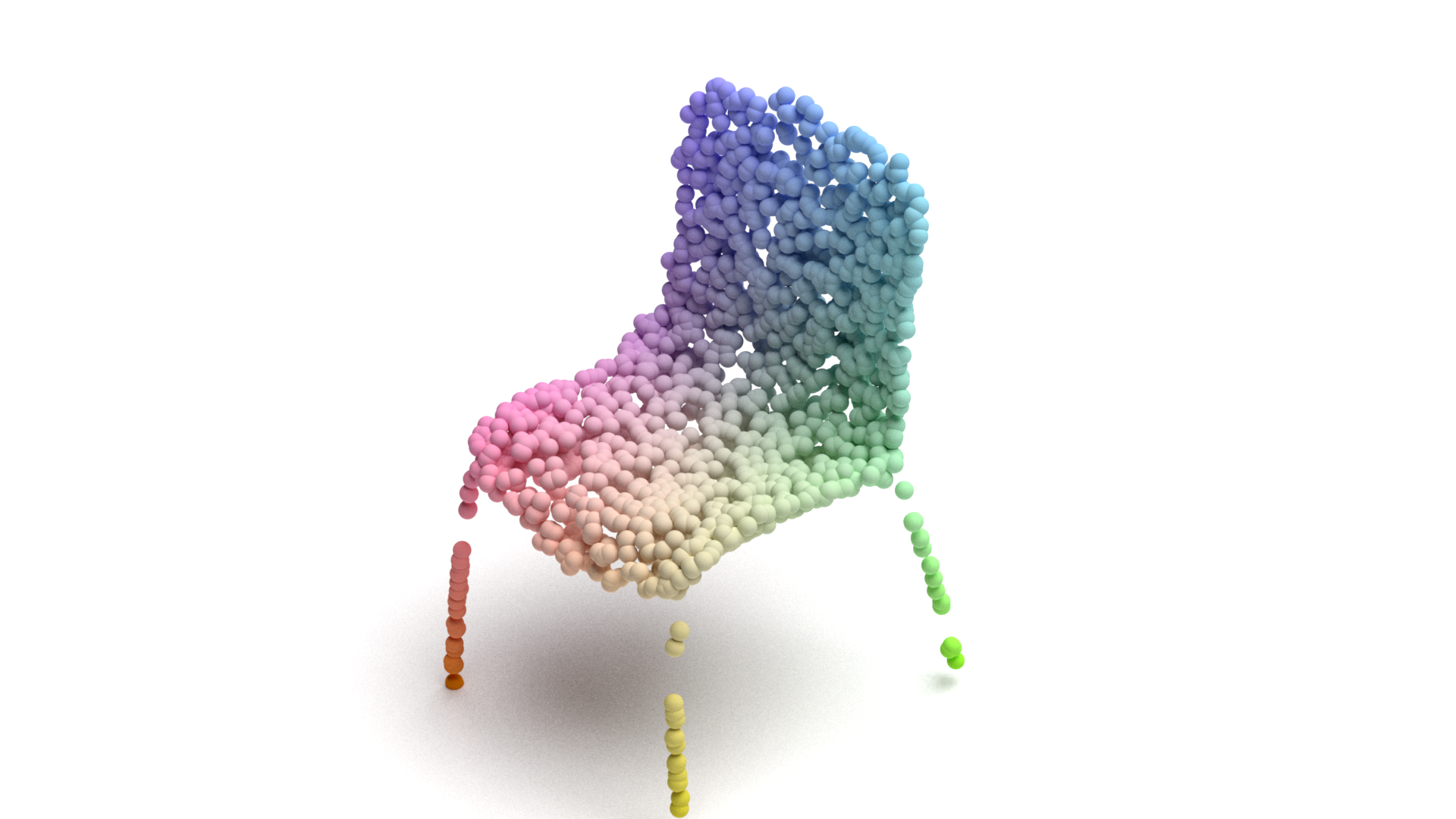} &
        \includegraphics[trim={15cm 0.0cm 15cm 0.0cm},clip, width=0.15\textwidth]{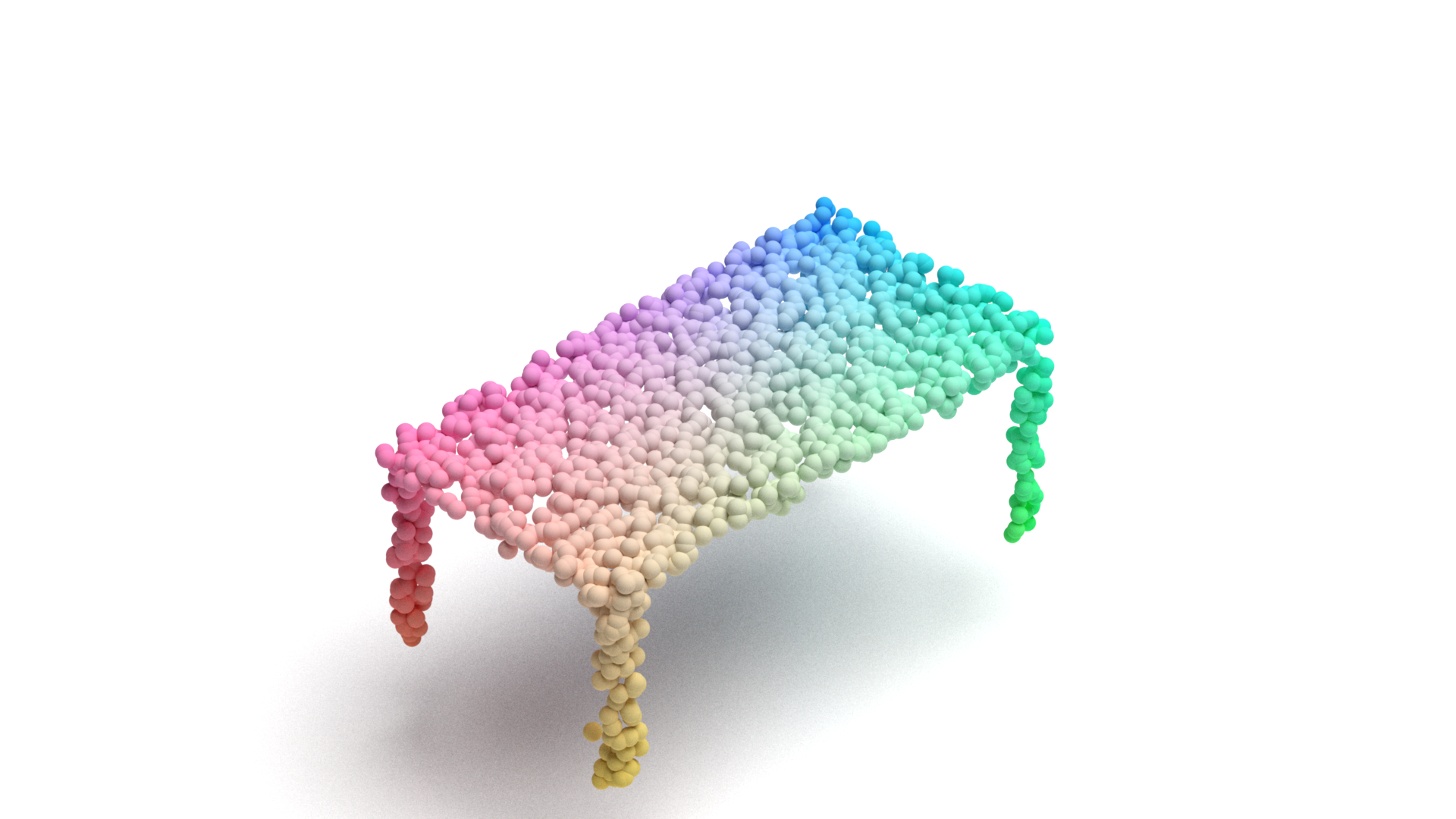} &
        \includegraphics[trim={15cm 0.0cm 15cm 0.0cm},clip,width=0.15\textwidth]{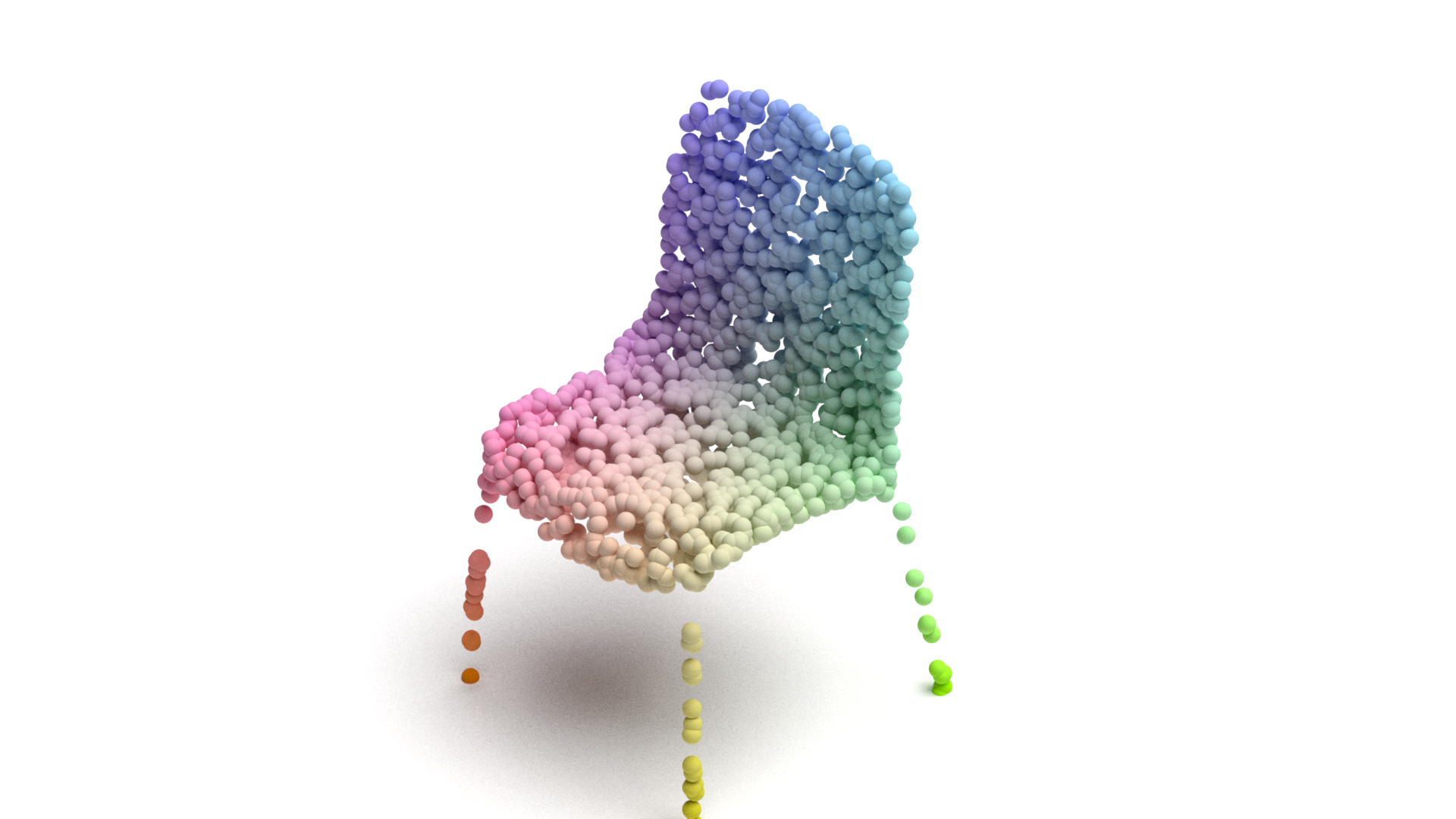} &
        \includegraphics[trim={15cm 0.0cm 15cm 0.0cm},clip,width=0.15\textwidth]{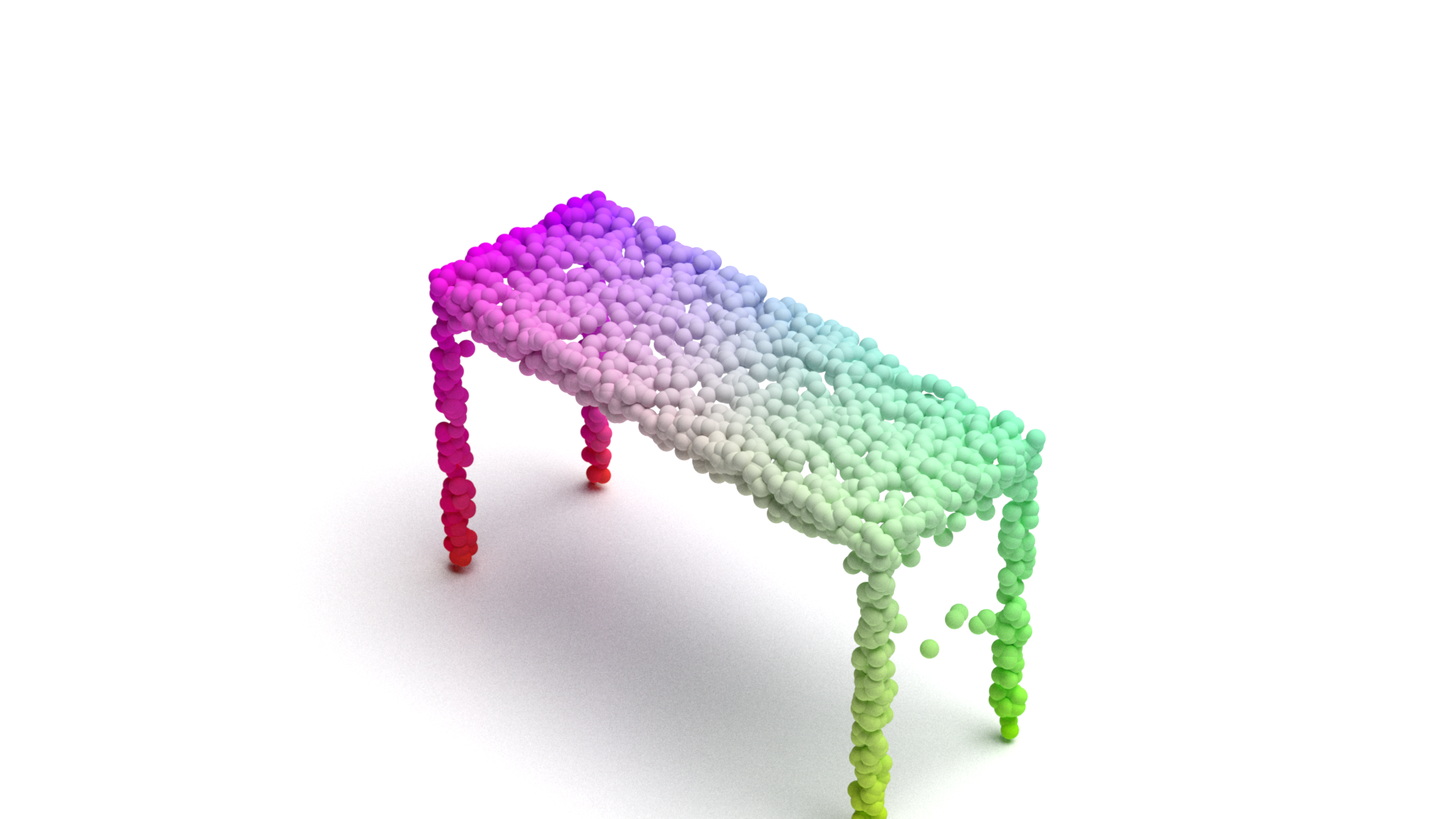} \\
        \includegraphics[trim={15cm 0.0cm 15cm 0.0cm},clip, width=0.15\textwidth]{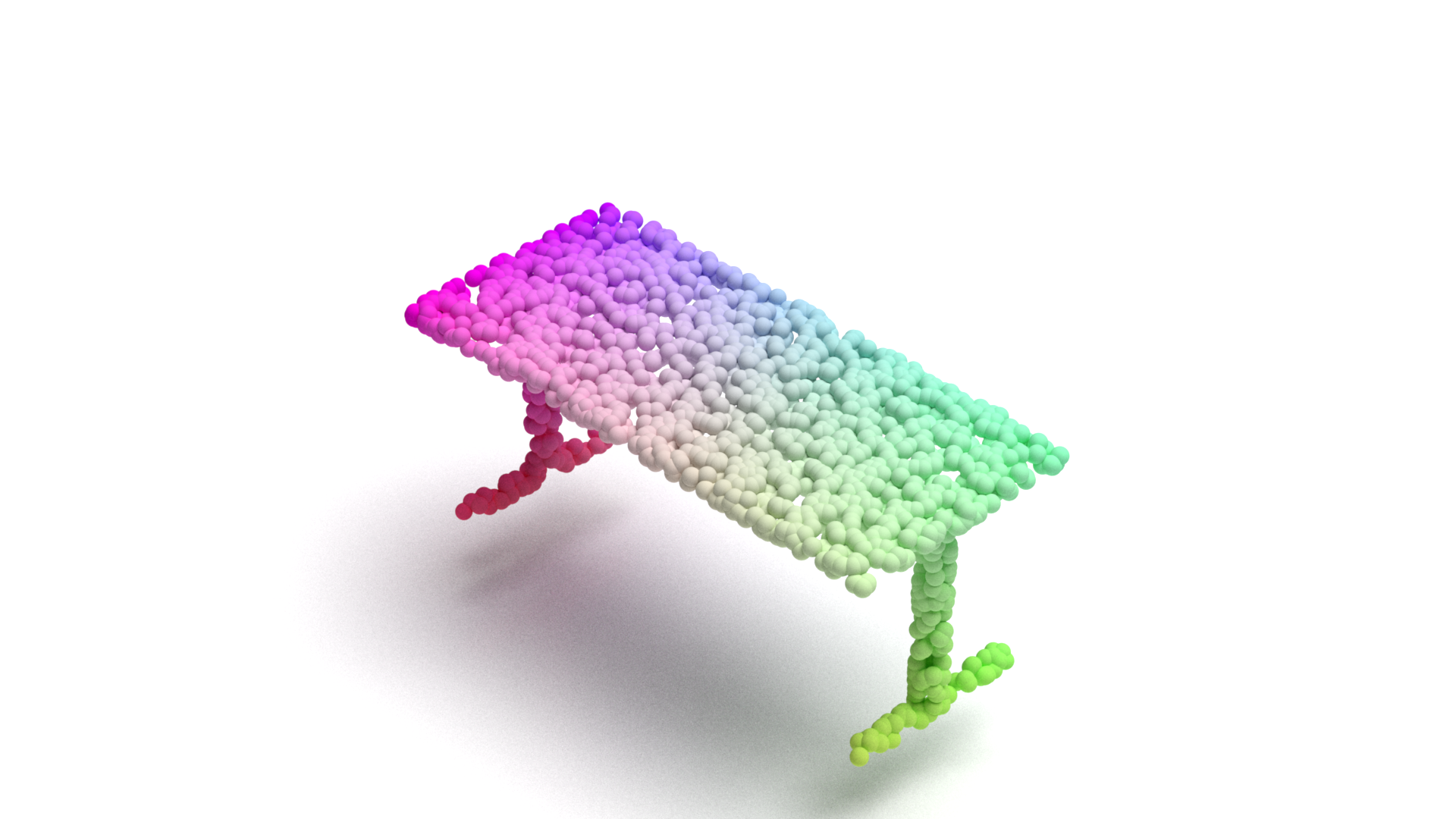} &
        \includegraphics[trim={15cm 0.0cm 15cm 0.0cm},clip, width=0.15\textwidth]{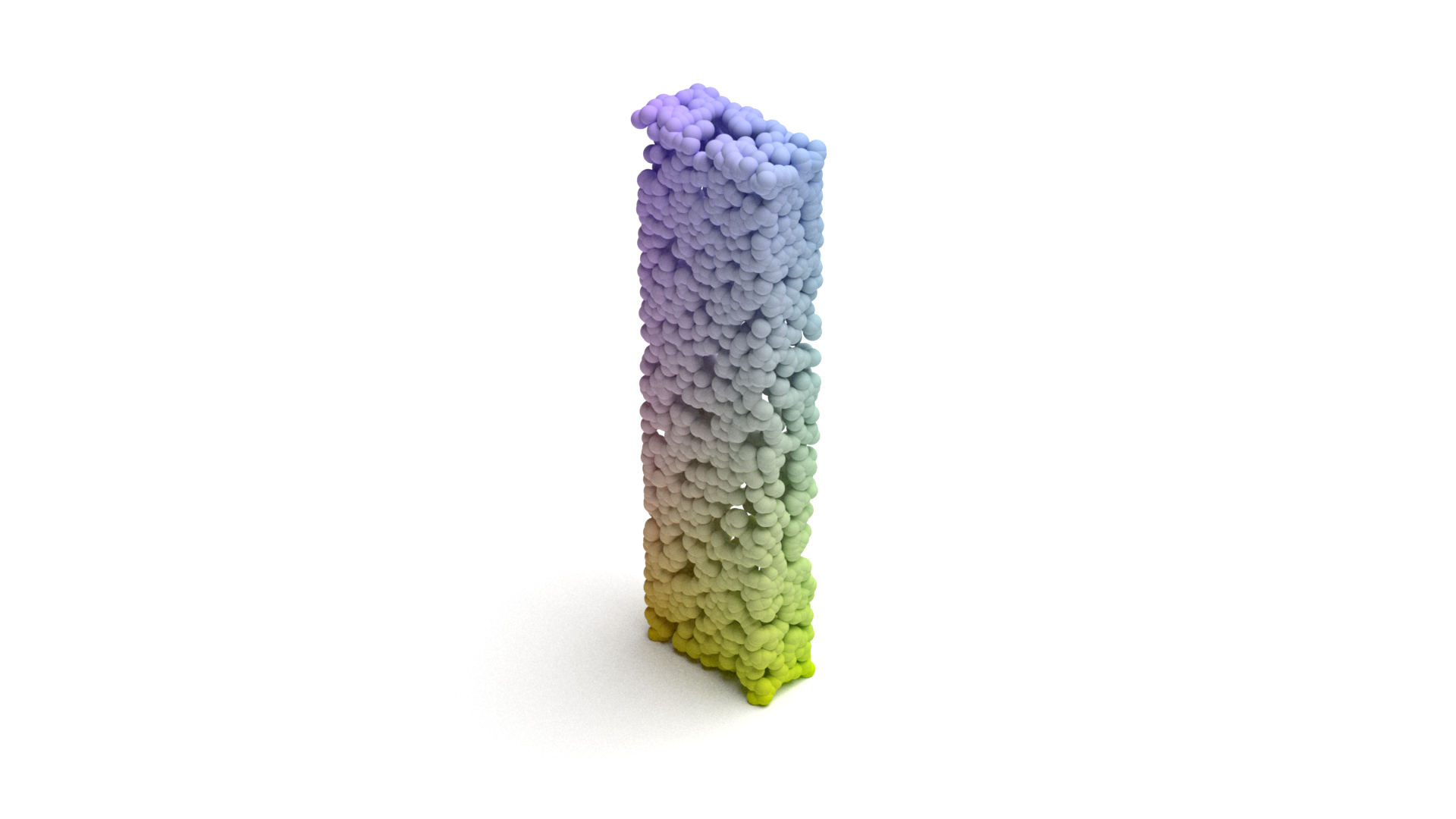} &
        \includegraphics[trim={15cm 0.0cm 15cm 0.0cm},clip, width=0.15\textwidth]{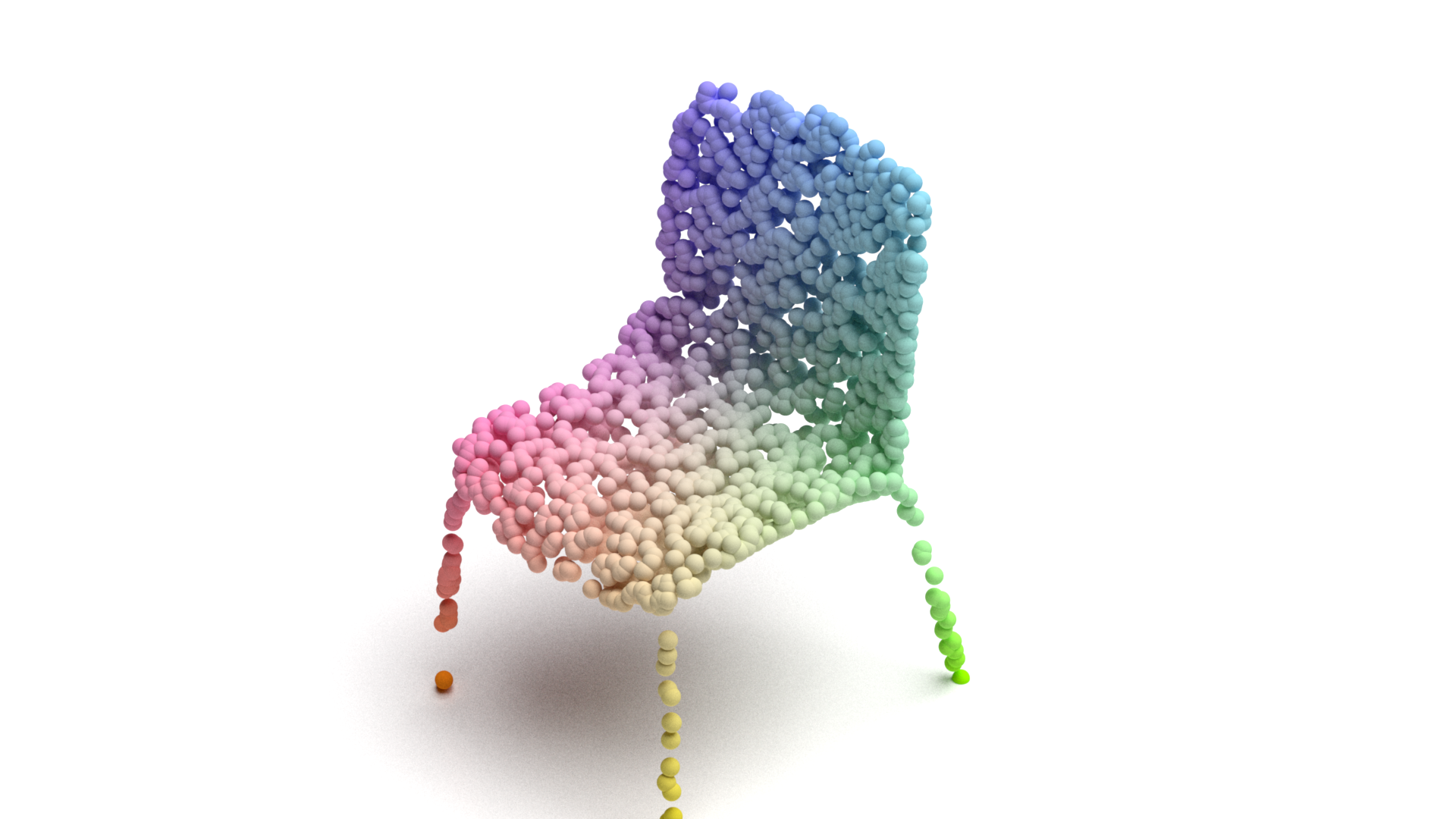} &
        \includegraphics[trim={15cm 0.0cm 15cm 0.0cm},clip, width=0.15\textwidth]{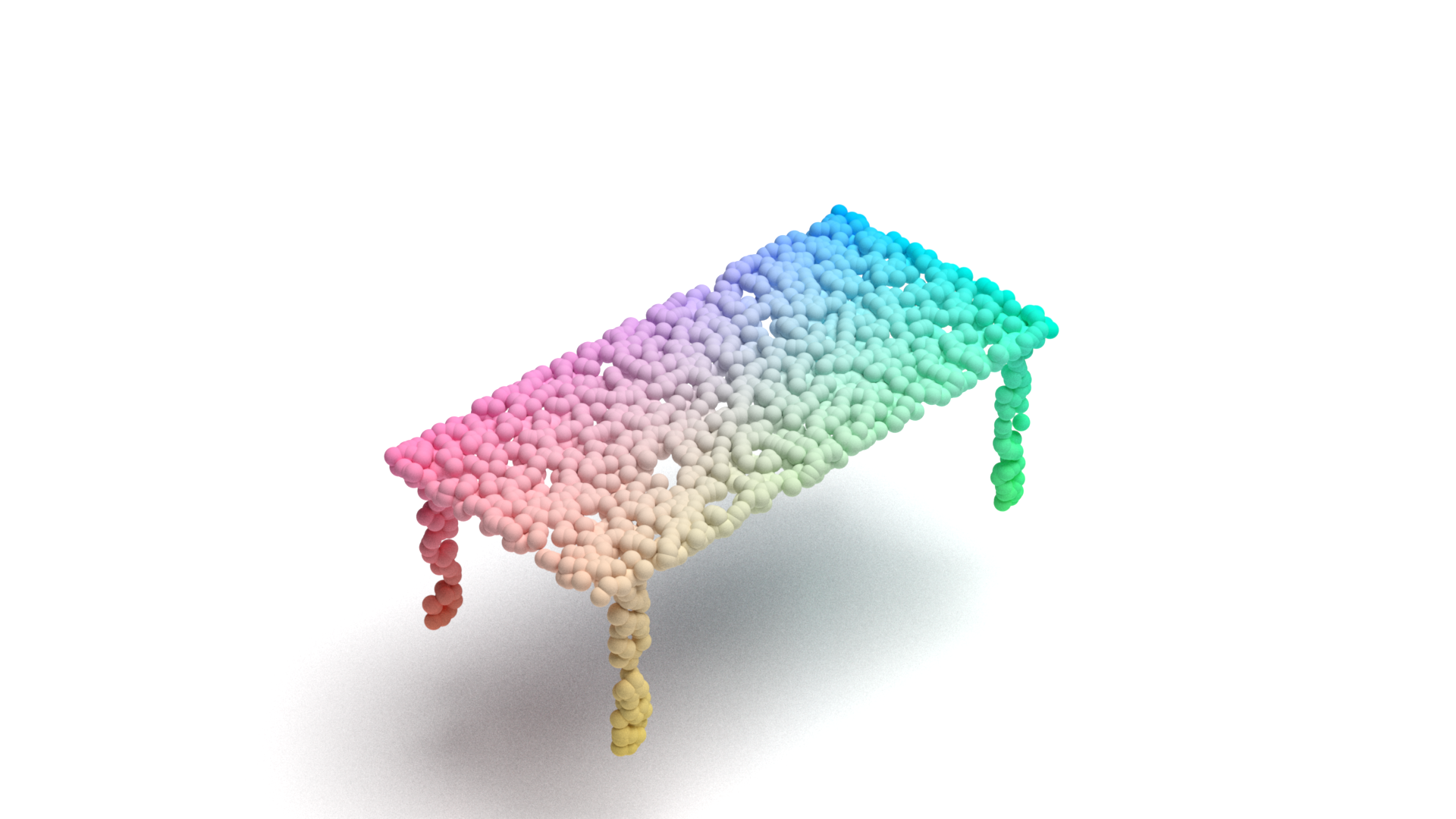} &
        \includegraphics[trim={15cm 0.0cm 15cm 0.0cm},clip,width=0.15\textwidth]{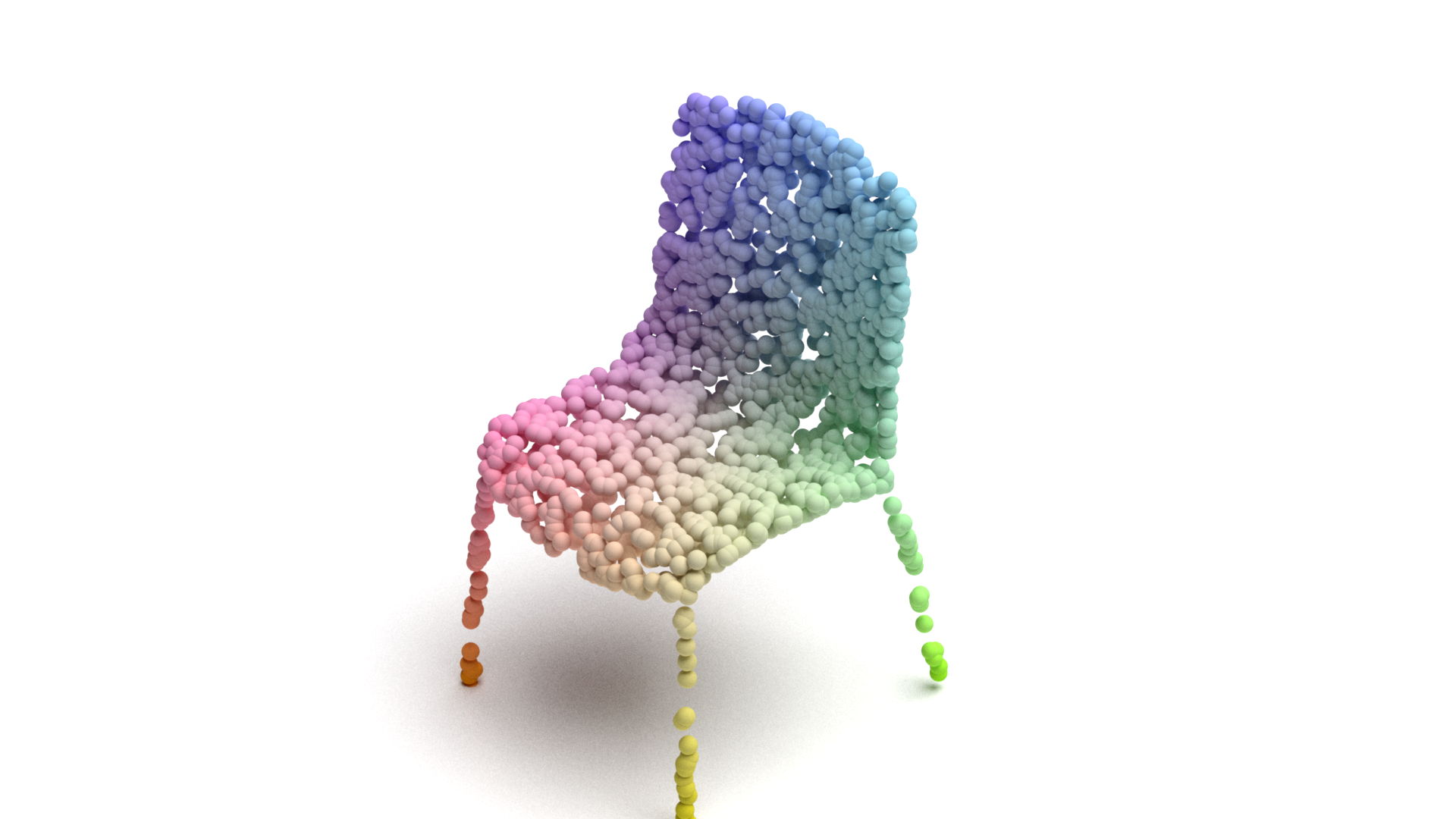} &
        \includegraphics[trim={15cm 0.0cm 15cm 0.0cm},clip,width=0.15\textwidth]{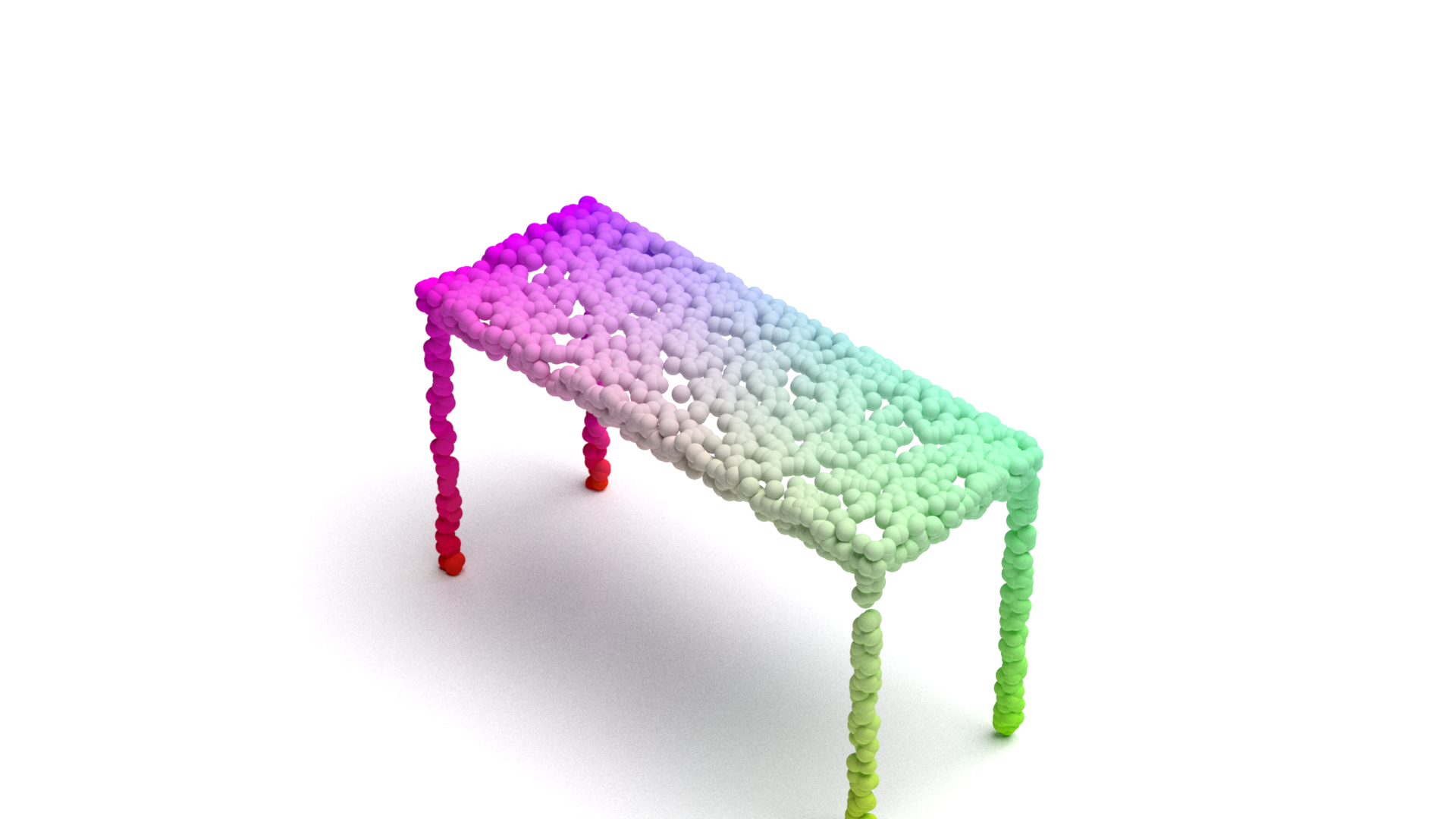} \\   
        \end{tabular}
        }
    \caption{\textbf{Qualitative results for point cloud completion on the SCANnotate++ dataset.} The visualization presents results in the following order: Top row shows partial point cloud input, middle row shows the reconstructed point cloud, and bottom row shows the ground truth point cloud. These results demonstrate the usefulness of our annotations for learning point cloud completion for ScanNet++~\cite{yeshwanth2023scannetpp}.} 
    \label{fig:result}
    \vspace{-3mm}
\end{figure*}

\begin{table*}
\centering
\scalebox{.9}{
\begin{tabular}{ccccccc}
\toprule
Dataset & CAD Annotations & \multicolumn{2}{c}{Alignment Accuracy $\uparrow$ } && \multicolumn{2}{c}{Retrieval-Aware Alignment Accuracy $\uparrow$} \\
\cmidrule{3-4} \cmidrule{6-7}
&& per class & per instance && per class & per instance \\
\hline
ScanNet &
Scan2CAD~\cite{avetisyan2019scan2cad} & 19.6  & 25.4 && 16.2 & 19.7  \\
ScanNet & SCANnotate~\cite{ainetter2023automatically,ainetter2024hoc} & \textbf{27.5} & \textbf{33.0} && \textbf{20.2} & \textbf{22.5} \\
\hline
ScanNet++ & SCANnotate++ (ours) & 27.3 & 33.2 && 17.7 & 13.8 \\
ScanNet++ & SCANnotate++ (ours)($^*$) & \textbf{30.5} & \textbf{34.9} && \textbf{20.6} & \textbf{16.1} \\
\bottomrule
\end{tabular}
}
\caption{\textbf{CAD model retrieval and alignment results for ROCA~\cite{gumeli2022roca} on different datasets.} Training on the automatic annotations from SCANnotate \cite{ainetter2023automatically,ainetter2024hoc} enhances performance on ScanNet compared to using manual annotations from Scan2CAD, proving highly effective for CAD model retrieval and alignment tasks. Similarly, our SCANnotate++ annotations enable generalization to ScanNet++. ($^*$) Pre-training on SCANnotate further boosts performance on ScanNet++. We use the official train and test splits from  ScanNet and ScanNet++.}
\label{tab:roca}
\vspace{-3mm}
\end{table*}

\begin{figure*}
    \centering
    \scalebox{.8}{
    \begin{tabular}{ccc||ccc}
        \includegraphics[width=0.18\textwidth]{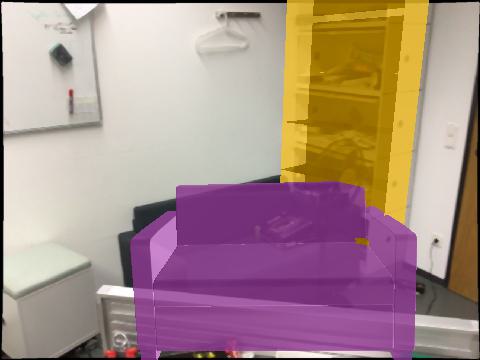} &
        \includegraphics[width=0.18\textwidth]{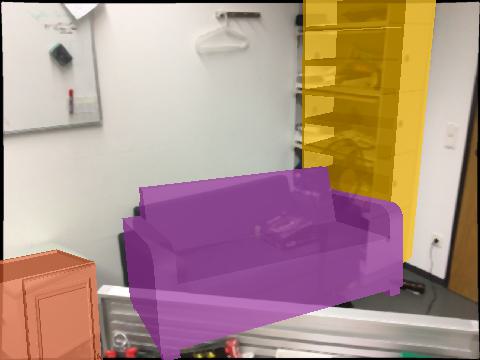} &
        \includegraphics[width=0.18\textwidth]{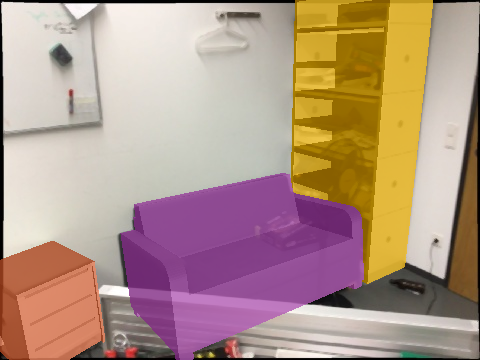} &
        \includegraphics[width=0.18\textwidth]{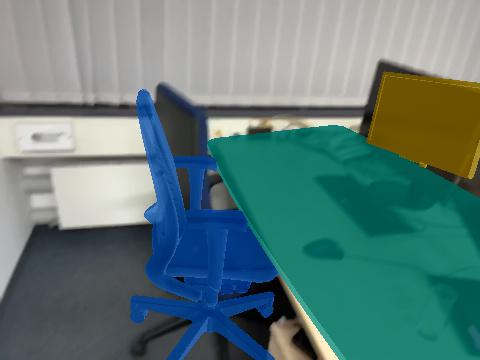} &
        \includegraphics[width=0.18\textwidth]{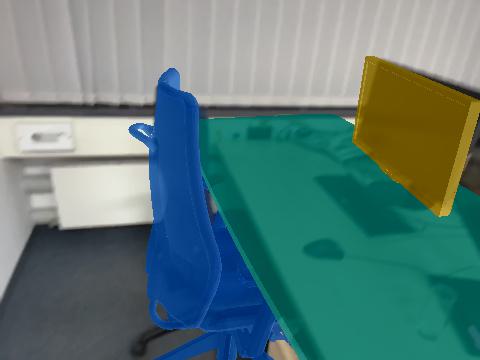} &
        \includegraphics[width=0.18\textwidth]{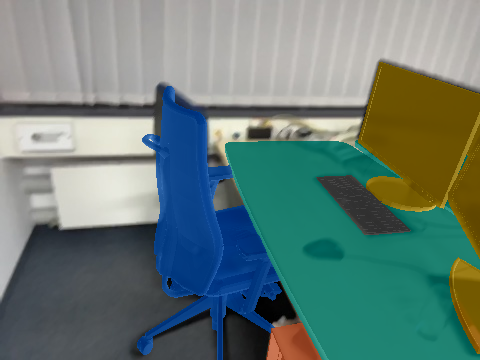} \\
        \includegraphics[width=0.18\textwidth]{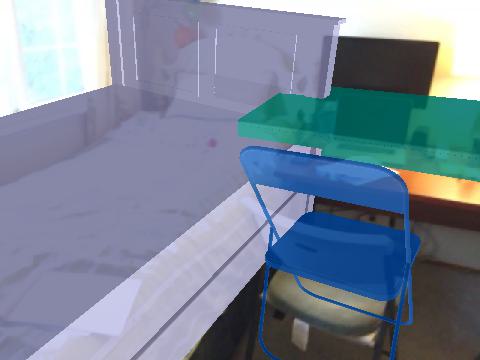} &
        \includegraphics[width=0.18\textwidth]{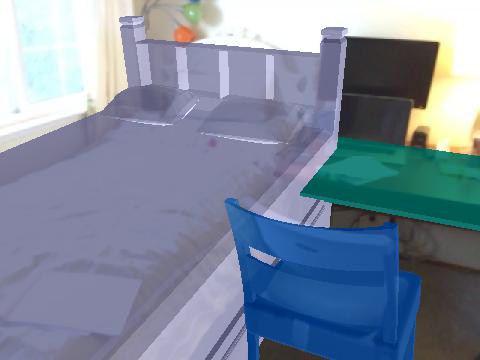} &
        \includegraphics[width=0.18\textwidth]{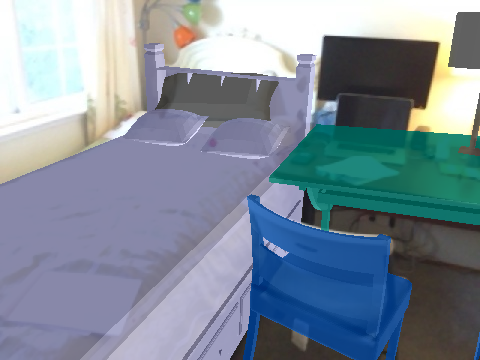} &
        \includegraphics[width=0.18\textwidth]{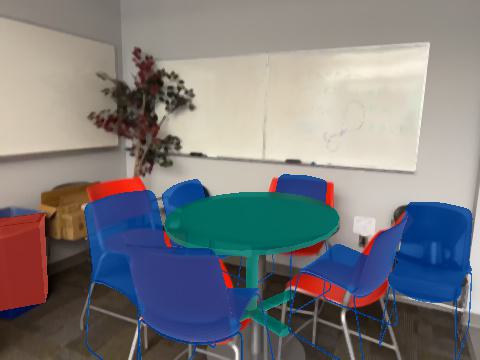} &
        \includegraphics[width=0.18\textwidth]{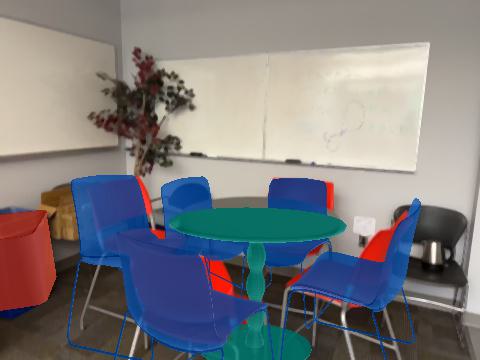} &
        \includegraphics[width=0.18\textwidth]{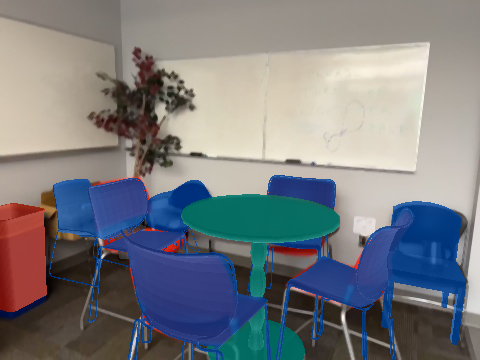} \\
        \includegraphics[width=0.18\textwidth]{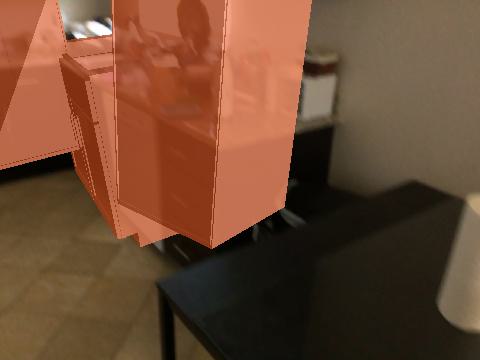} &
        \includegraphics[width=0.18\textwidth]{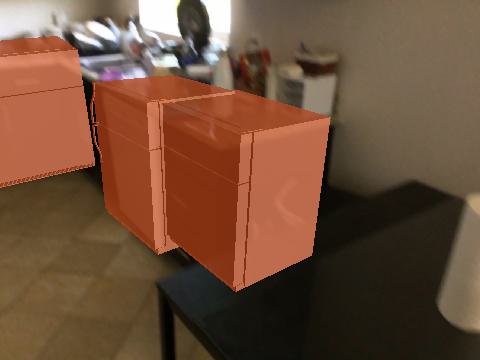} &
        \includegraphics[width=0.18\textwidth]{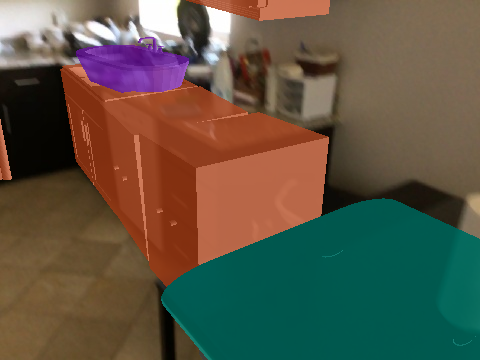} &
        \includegraphics[width=0.18\textwidth]{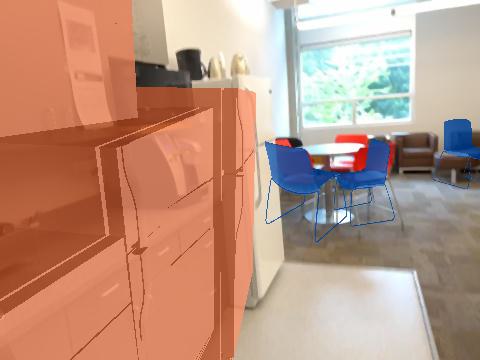} &
        \includegraphics[width=0.18\textwidth]{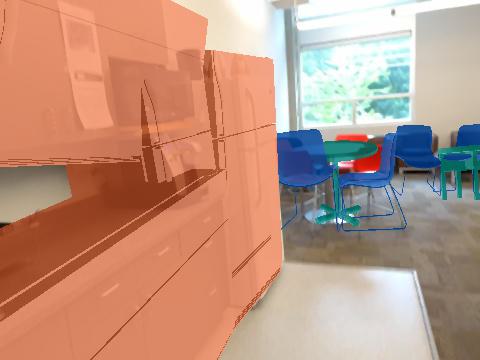} &
        \includegraphics[width=0.18\textwidth]{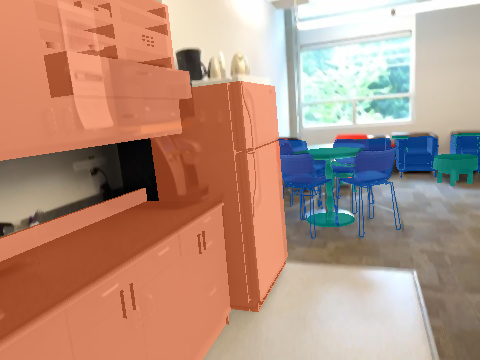} \\
        \includegraphics[width=0.18\textwidth]{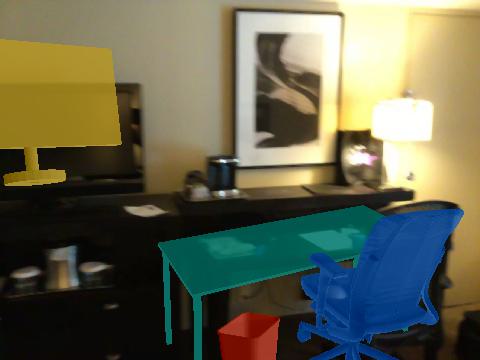} &
        \includegraphics[width=0.18\textwidth]{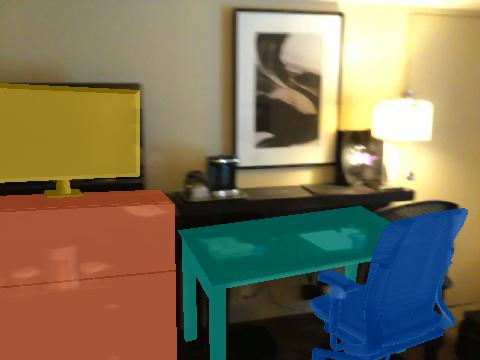} &
        \includegraphics[width=0.18\textwidth]{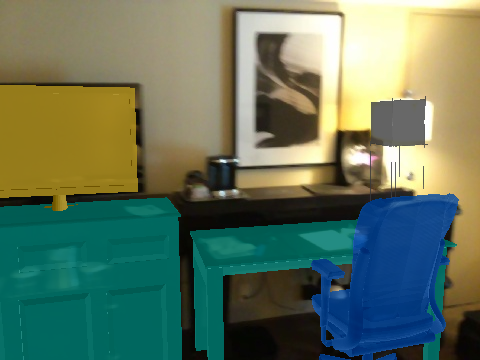} &
        \includegraphics[width=0.18\textwidth]{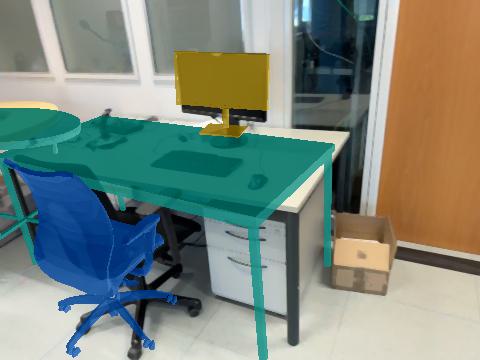} &
        \includegraphics[width=0.18\textwidth]{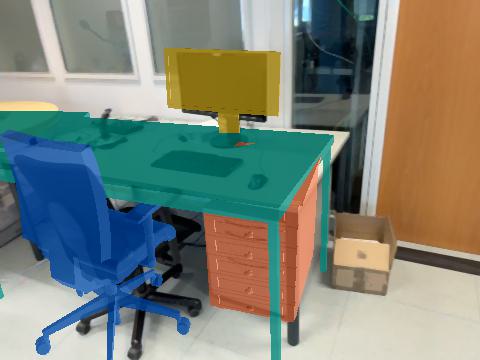} &
        \includegraphics[width=0.18\textwidth]{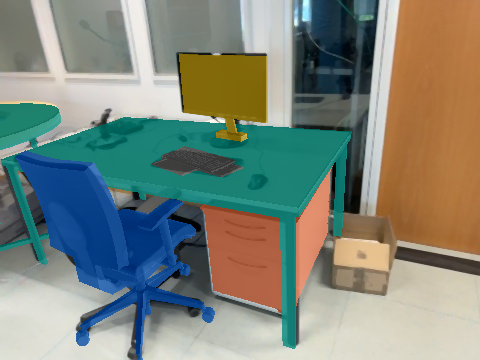} \\
        Scan2CAD & SCANnotate & SCANotate g.t. & SCANnotate++ & ($^*$) SCANnotate++ & SCANnotate++ g.t. \\
    \end{tabular}
    }
    \caption{\textbf{Qualitative results for CAD model retrieval and alignment for ROCA~\cite{gumeli2022roca} trained on different annotations.}  Training ROCA on automatic annotations from SCANnotate~\cite{ainetter2023automatically,ainetter2024hoc} demonstrates improved alignment of retrieved CAD models over manual annotations from Scan2CAD~\cite{avetisyan2019scan2cad}. Training on SCANnotate++ enables good performance on ScanNet++~\cite{yeshwanth2023scannetpp}, and pre-training on SCANnotate~($^*$) helps to further improve these results.} 
    \label{fig:roca}
    \vspace{-0.3cm}
\end{figure*}

\paragraph{Training details.} To generate the input point cloud, we first extract partial object meshes from the target objects in  the ScanNet~\cite{dai2017scannet} or ScanNet++~\cite{yeshwanth2023scannetpp} datasets and uniformly sample $2048$ points from these meshes. We obtain the ground truth point clouds by directly sampling $2048$ points from the corresponding ground truth meshes. During training, we followed the ShapeGF configuration by using the Adam optimizer with a batch size of 32, a learning rate of 0.001, and the training for 1200 epochs.

\paragraph{Evaluation.} We evaluate our completion method on test sets of manual annotations from Scan2CAD~\cite{avetisyan2019scan2cad}, and automatic annotations from SCANnotate~\cite{ainetter2023automatically,ainetter2024hoc} and SCANnotate++. We measure the performance using the Chamfer Distance~(CD) and Earth Mover's Distance~(EMD).

As Table~\ref{tab:completion} shows, our point cloud completion method can be trained on automatically generated training data to produce high-quality results. To further confirm the quality of automatic annotations, we provide results of training our method on manual annotations from Scan2CAD~\cite{avetisyan2019scan2cad}. Even when tested on the test set of Scan2CAD annotations, the method when using our dataset performs much better. We argue that SCANnotate contains more annotations than Scan2CAD, which naturally leads to better performance. Increasing the number of automatic annotations by including our SCANnotate++ annotations improves the quality of reconstructions even further. Nonetheless, due to the limited size of ScanNet++~v1 and SCANnotate++, we were not able to train our model solely on ScanNet++~v1, and hence we exclude these results. We also evaluated the performance of SCANnotate and SCANnotate++ using only an auto-encoder, which yielded a CD of 210.35 and an EMD of 20.20. Figure~\ref{fig:result} shows additional qualitative results for point cloud completion on our SCANnotate++ dataset.

\subsection{CAD Model Retrieval and Alignment}

Given a single image, we want to retrieve and align corresponding 3D CAD models. ROCA~\cite{gumeli2022roca} showed that supervised learning methods can be applied when trained on manual annotations from Scan2CAD~\cite{avetisyan2019scan2cad}. We show such methods can be effectively trained on automatic annotations from SCANnotate~\cite{ainetter2023automatically,ainetter2024hoc} and our SCANnotate++. 

\paragraph{ROCA~\cite{gumeli2022roca}.}
ROCA is a framework that, from a single image, retrieves and aligns CAD models from a pool of per-scene objects. It identifies a CAD model that closely resembles the target object in both appearance and geometry, while concurrently estimating 9-DoF alignment. This is achieved through a differentiable optimization process that leverages dense 2D-3D correspondences alongside Procrustes alignment.

\paragraph{Dataset processing.} Alignment of objects can sometimes be ambiguous due to object symmetries. Therefore, similarly to Scan2CAD, we extract symmetries for the target objects. To determine the symmetry level of a CAD model, we rotate the original mesh in increments of 45 degrees. For each rotation, we uniformly sample $5000$ points from both the original and rotated meshes, and then compute the Chamfer Distance between them. A threshold of $0.05$ is used to determine whether the two models are considered identical or not. Based on the range of valid rotation angles, we categorize the models into four rotation groups, following the settings defined in Scan2CAD~\cite{avetisyan2019scan2cad}.

\paragraph{Evaluation.} As in our first application, we evaluate our proposed method on the test sets of manual annotations from Scan2CAD~\cite{avetisyan2019scan2cad}, and automatic annotations from SCANnotate~\cite{ainetter2023automatically,ainetter2024hoc}, and our SCANnotate++. In line with the methodology described in ROCA, we evaluate the performance using two metrics: \textit{Alignment Accuracy} for 9-DoF alignment and \textit{Retrieval-Aware Alignment Accuracy} for retrieval. An \textit{Alignment} is 'correct' if the object classification is accurate and the translation error is within $20cm$, the rotation error is within $20^\circ$, and the scale ratio error is within $20\%$. A \textit{Retrieval-Aware Alignment} is considered correct if the Alignment is correct and the retrieved CAD model class matches the ground truth class.

Table~\ref{tab:roca} presents class-averaged and instance-averaged performance results for Alignment Accuracy and Retrieval-Aware Alignment Accuracy across the three datasets. In comparison to Scan2CAD~\cite{avetisyan2019scan2cad}, automatic SCANnotate~\cite{ainetter2023automatically} annotations yield better CAD model and 9-DoF alignments for the ScanNet dataset~\cite{dai2017scannet}. Additionally, utilizing a model pretrained  on SCANnotate~\cite{ainetter2023automatically,ainetter2024hoc} before training on our SCANnotate++ annotations, improves overall performance on SCANnotate++, demonstrating that increasing the number of automatic annotations further benefits the training. Qualitative results in Figure~\ref{fig:roca} further reinforce these conclusions: ROCA~\cite{gumeli2022roca} achieves better CAD model alignment and retrieval results with SCANnotate annotations~\cite{ainetter2023automatically,ainetter2024hoc} compared to Scan2CAD~\cite{avetisyan2019scan2cad}, while training on both SCANnotate and SCANnotate++ further improves alignments.

\section{Conclusion}

We integrated recent methods for automatic object shape and pose estimation into an end-to-end pipeline, generating high-quality annotations for unannotated datasets like ScanNet++. We demonstrated the benefits of these annotations for training learning-based models and believe that leveraging more data thanks to such a pipeline can significantly boost the performance of current 3D deep learning methods.


\section*{Acknowledgments}
This project was funded in part by the European Union (ERC Advanced Grant explorer Funding ID \#101097259).

{
    \small
    \bibliographystyle{ieeenat_fullname}
    \bibliography{main}
}


\end{document}